\documentclass[sn-mathphys,Numbered,iicol]{sn-jnl}

\usepackage{graphicx}%
\usepackage{array}%
\usepackage{multirow}%
\usepackage{amsmath,amssymb,amsfonts}%
\usepackage{amsthm}%
\usepackage{mathrsfs}%
\usepackage[title]{appendix}%
\usepackage{xcolor}%
\usepackage{textcomp}%
\usepackage{manyfoot}%
\usepackage{booktabs}%
\usepackage{algorithm}%
\usepackage{algorithmicx}%
\usepackage{algpseudocode}%
\usepackage{listings}%

\theoremstyle{thmstyleone}%
\newtheorem{theorem}{Theorem}
\newtheorem{proposition}[theorem]{Proposition}%

\theoremstyle{thmstyletwo}%
\newtheorem{remark}{Remark}%
\newcommand{\cart}{\ensuremath{\raisebox{-0.5mm}{\mbox{\LARGE{$\times$}}}}}

\theoremstyle{thmstylethree}%
\newtheorem{definition}{Definition}%

\usepackage{xr}
\makeatletter
\newcommand*{\addFileDependency}[1]{
\typeout{(#1)}
\@addtofilelist{#1}
\IfFileExists{#1}{}{\typeout{No file #1.}}
}\makeatother

\newcommand*{\myexternaldocument}[1]{%
\externaldocument[supp-]{#1}%
\addFileDependency{#1.tex}%
\addFileDependency{#1.aux}%
}
\myexternaldocument{AAAA_SSVM_SUPPLEMENTARY_MATERIAL}

\newtheorem{lemma}{Lemma}

\usepackage{amsopn}

\usepackage{cancel} 
\usepackage{soul}
\usepackage{makecell}
\usepackage{caption}
\usepackage{multirow}

\newcommand{\Rm}{\mathbb{R}}
\newcommand{\Rmn}{\mathbb{R}^n}
\newcommand{\GGD}{\mathcal{GGD}}

\newtheorem{assumption}{Assumption}
\usepackage{enumerate}
\usepackage[T5,T1]{fontenc}

\newcommand{\fonc}[3]{#1:  #2  \longrightarrow  #3}					
\newcommand{\syst}[1]{\left \{ \begin{array}{l} #1 \end{array} \right. \kern-\nulldelimiterspace}	
\newcommand{\prox}{\text{\normalfont prox}}
\newcommand{\argmin}{\text{\normalfont argmin}}

\newcommand{\dom}{\text{\normalfont dom}\,}

\newcommand{\minimize}[2]{\ensuremath{\underset{\substack{{#1}}}{\mathrm{minimize}}\;\;#2 }}

\newcommand{\argmind}[2]{\ensuremath{\underset{\substack{{#1}}}%
{\mathrm{argmin}}\;\;#2 }}

\newcommand{\N}{\mathbb{N}}		
\newcommand{\R}{\mathbb{R}}		

\usepackage{pdfpages} 
\usepackage{pgffor} 

\makeatletter
\AtBeginDocument{\let\LS@rot\@undefined}
\makeatother

\def\supplementfilename{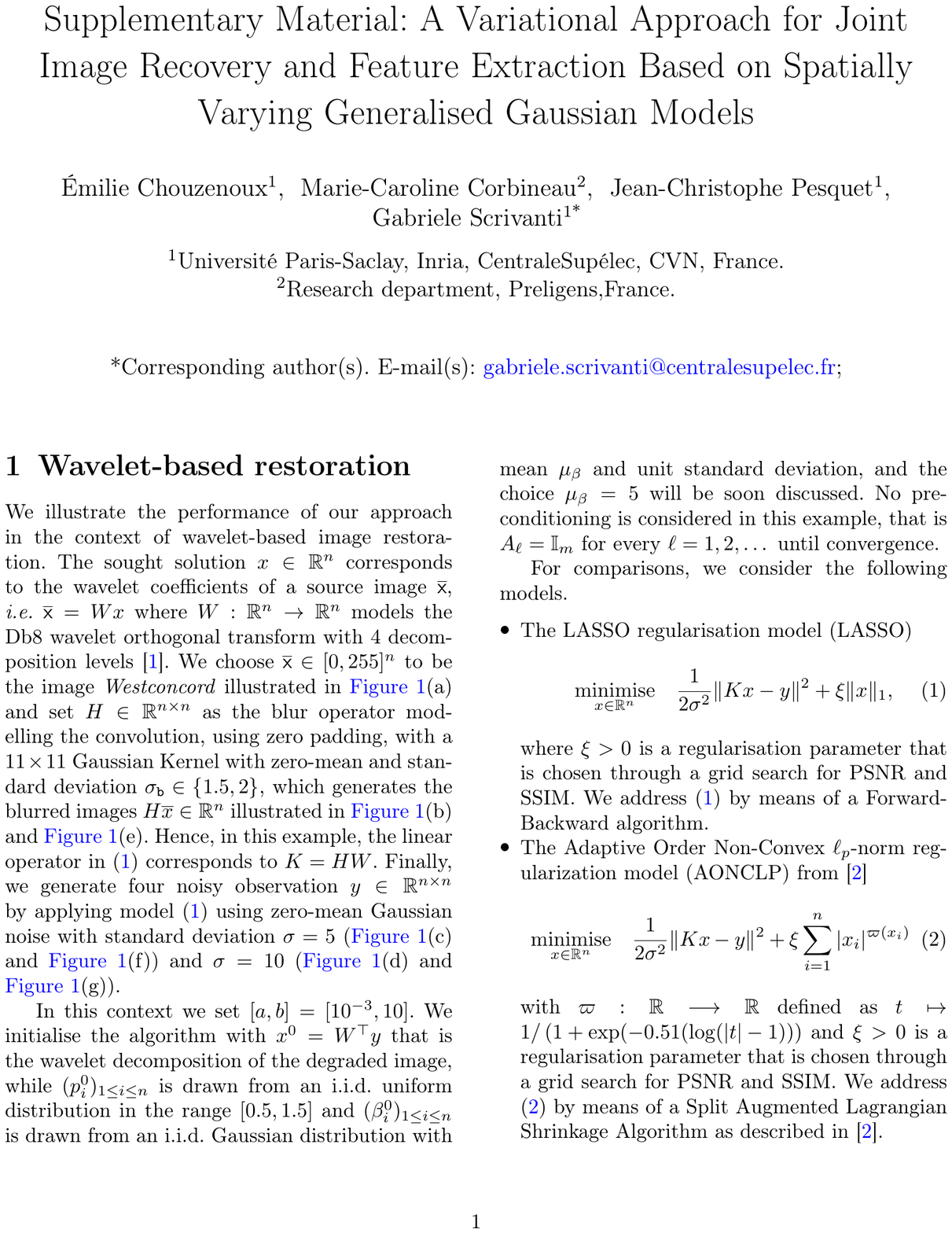}

\pdfximage{\supplementfilename}
\def\numbersupplementpages{\the\pdflastximagepages}

\newif\ifarXiv
\arXivtrue 

\begin{document}

\title[Title]{{
A Variational Approach for Joint Image Recovery and Feature Extraction
Based on Spatially Varying Generalised Gaussian Models}}

\author[1]{\fnm{\'Emilie} \sur{Chouzenoux}}
\author[2]{\fnm{Marie-Caroline} \sur{Corbineau}}
\author[1]{\fnm{Jean-Christophe} \sur{Pesquet}}
\author*[1]{\fnm{Gabriele} \sur{Scrivanti}}\email{gabriele.scrivanti@centralesupelec.fr}

\affil[1]{\orgname{Université Paris-Saclay, Inria, CentraleSupélec, CVN}, \state{France}}
\affil[2]{\orgname{Research department, Preligens},\state{France}}


\abstract{
The joint problem of reconstruction/feature extraction is a challenging task in image processing. It consists in performing, in a joint manner, the restoration of an image and the extraction of its features. In this work, we firstly propose a novel non-smooth and non-convex variational formulation of the problem. For this purpose, we introduce a versatile generalised Gaussian prior whose parameters, including its exponent, are space-variant.
Secondly, we design an alternating proximal-based optimisation algorithm that efficiently exploits the structure of the proposed non-convex objective function. We also analyse the convergence of this algorithm. As shown in numerical experiments conducted on joint deblurring/segmentation tasks, the proposed method provides high-quality results.}  

\keywords{
Image recovery ; Space-variant regularisation ; Alternating minimization ; Proximal algorithm ; Block coordinate descent ; 
Image segmentation 
}

\maketitle

\section{Introduction}

Variational regularisation of ill-posed inverse problems in ima\-ging relies on the idea of searching for a solution in a well-suited space. A central role in this context is played by $\ell_p$ spaces with $p\in (0,\infty)$, and the power $p$ of the corresponding norms when $p\geq 1$ \cite{Daubechies,Grasmair2008SparseRW,Lorenz+2008+463+478,RamlauResmerita,LASSO} or seminorms when $p\in(0,1)$ \cite{Chartrand,Grasmair,Zarzer2009OnTR}. {For every vector $u=(u_i)_{1\le i \le n}\in \mathbb{R}^n$ and $p\in(0,+\infty)$, the $\ell_p$  (semi-)norm is denoted by $
    \|u\|_p  = \big(\sum_{i=1}^{n} |u_i|^p \big) ^ {1/p}$. We usually omit $p$ when $p=2$, so that $\|\cdot \| = \|  \cdot \|_2$. The case $p\in (0,1)$} has gained rising credit, especially in the field of sparse regularisation. An extensive literature has been focused on challenging numerical tasks raised by the non-convexity of the seminorms and the possibility of combining them with linear operators to extract salient features of the sought images \cite{Ghilli2019OnMA,Hintermller2013NonconvexTI}. In \cite{lorenzresmerita}, the more general notion of $F$-norm is introduced in order to establish functional analysis results on products of $\ell_{p_i}$-spaces with $p_i \in (0,2]$. For some $x = (x_i)_{1 \leq i \leq n}\in \mathbb{R}^n$, this amounts to studying the properties of penalties of the form $\sum_{i=1}^n |x_i|^{p_i}$, for some positive exponents $(p_i)_{1\le i \le n}$. This approach offers a more flexible framework by considering a wider range of exponents than the standard $\ell_p$-based regularisation. However, it extends the problem of choosing a suitable exponent $p$ to a whole sequence of exponents $(p_i)_{1\leq i \leq n}$. {  
 In \cite{Afonso_2017}, the authors proposed a non-convex regulariser of the form $\sum_{i=1}^n |x_i|^{\varpi(|x_i|)}$, where each exponent is expressed as a function of the absolute magnitude of the data and function $\varpi(\cdot)$ is a rescaled version of the sigmoid function, taking values in the interval $[0,1]$.} In image restoration, a similar approach consists in adopting space variant regularisation models built around a Total Variation-like functional with a variable exponent $\sum_{i=1}^n \|(\nabla x)_i\|^{p_i}$ where $\nabla$ is a discrete 2D gradient operator. The rationale is to select the set of parameters $(p_i)_{1\leq i \leq n}$ in order to promote either edge enhancement ($p_i=1$) or smoothing ($p_i>1$) depending on the spatial location encoded by index $i$. This model was introduced in \cite{blomgren} and then put into practice firstly for $p_i\in [1,2]$ in \cite{Chen2006VariableEL} and then for $p_i \in(0,2] $  in \cite{SVTVP}. {  To conclude, in a recent work \cite{Lazzaretti2022}, the authors proposed a modular-proximal gradient algorithm to find solutions to ill-posed inverse problems in variable exponents Lebesgue spaces $L^{p(\cdot)}(\Omega)$ with $\Omega\subseteq\R^n$, rather than in $L^{2}(\Omega)$.} In all of these works, the so-called space variant $p$-map (\emph{i.e.}\ , $(p_i)_{1\leq i \leq n}$) is estimated offline in a preliminary step 
and then kept fixed throughout the optimisation procedure. 

   
In this paper, we address the problem of joint image recovery and feature extraction. Image recovery amounts to retrieving an estimate of an original image from a degraded version of it. The degradation usually corresponds to the application of a linear operator (e.g., blur, projection matrix) to the image and the addition of a noise. Feature extraction problems arise when one wants to assign to an image a small set of parameters which can describe or identify the image itself. Image segmentation can be viewed as an example of feature extraction, which consists of defining a label field on the image domain so that pixels are partitioned into a predefined number of homogeneous regions according to some specific characteristics. {  A second example, similar to segmentation, is edge detection, where one aims at identifying the contour changes within different regions of the image. }Texture retrieval is a third example. This task relies on the idea of assigning a set of parameters to each coefficient of the image -- possibly in some transformed space -- so that the combination of all parameters defines a "signature" that represents the content of various spatial regions. Joint image recovery and feature extraction consists in performing, in a joint manner, the image recovery and the extraction of features in the sought image.  

A powerful and versatile approach for feature extraction, that we propose to adopt here, assumes that the data follow a mixture of generalised Gaussian probability distribution $(\mathcal{GGD})$~\cite{PPULA,WAVELET_GGD, US_GGD}. The $\mathcal{GGD}$ model results in a sum of weighted $\ell_{p_i}$-based terms in the criterion, with general form $\sum_{i=1}^n \vartheta_i | x_i|^{p_i}$ with $\{\vartheta_i\}_{1\le i \le n} \subset [0,+\infty)$. We thus aim at jointly estimating an optimal configuration for $(\vartheta_i,p_i)_{1 \leq i \leq n}$, and retrieving the image. Under an assumption of consistency within the exponents' values of a given region of the features space, we indeed obtain the desired feature extraction starting from the estimated $p$-map. The latter amounts to minimizing a non-smooth and non-convex cost function.

This specific structure of the proposed objective function suggests the use of an alternating minimisation procedure. In such an approach, one sequentially updates a subset of parameters through the resolution of an inner minimization problem, while the other parameters are assumed to be fixed. This approach has a standard form in the Block Coordinate Descent method (BCD) (also known as Gauss-Seidel algorithm) \cite{HILDRETH}. In the context of non-smooth and non-convex problems, the simple BCD may, however, show instabilities \cite{Tseng2001ConvergenceOA}, which resulted in an extensive construction of alternative methods that efficiently exploit the characteristics of the functions, and introduce powerful tools to improve the convergence guarantees of BCD, or overcome difficulties arising in some formulations.{ In this respect, a central role is played by proximal methods \cite{Combettes2011,FIXEDPOINTSTRATEGIES}:} a proximally regularised BCD (PAM) for non-convex problems was studied in \cite{PAM}; a proximal linearised method (PALM) and its inertial and stochastic versions 
were then proposed in \cite{PALM} resp. \cite{iPALM} and \cite{hertrich2020inertial}; in \cite{SLPAM}, the authors investigated the advantage of a hybrid semi-linearised scheme (SL-PAM) 
for the joint task of image restoration and edge detection based on a discrete version of the Mumford–Shah model. A structure-adapted version of PALM (ASAP) 
was designed in \cite{ASAP, Tan2019InertialAG} to exploit the block-convexity of the coupling terms and the regularity of the block-separable terms arising in some practical applications such as image colorisation and blind source separation. The extension to proximal mappings defined w.r.t. a variable metric was firstly introduced in \cite{Censor1987OptimizationO}, leading to the so-called Block Coordinate Variable Metric Forward-Backward. An Inexact version 
and a line search based version 
of it were presented in \cite{BCVMFB} and \cite{AVMILA}, respectively. In \cite{VMFB_COMPOSITE_MINIMISATION} the authors introduced a Majorisation-Minimisation (MM) strategy within a Variable Metric Forward-Backward algorithm to tackle the challenging task of computing the proximity operator of composite functions. We refer to \cite{Bonettini2019} for an in-depth analysis of how to introduce a variable metric into first-order methods.
{  In \cite{Hien2023titan}, the authors introduced a family of block-coordinate majorisation-minimisation methods named TITAN. Various majorisation strategies can be encompassed by their framework, such as proximal surrogates, Lipschitz gradient surrogates,  or Bregman surrogate functions. Convergence of the algorithm iterates are shown in \cite{Hien2023titan,BCVMFB}, under mild assumptions, that include the challenging non-convex setting. These studies emphasised the prominent role played by the Kurdyka-\L{}ojasiewicz (KL) inequality \cite{attouch-bolte-svaiter}.  

In the proposed problem formulation, the objective function includes several non-smooth terms, as well as a quadratic term -- hence Lipschitz differentiable -- that is restricted to a single block of variables. This feature makes the related subproblem well-suited for a splitting procedure that involves an explicit gradient step with respect to this term, combined with implicit proximal steps on the remaining blocks of variables. Variable metrics within gradient/proximal steps would also be desirable for convergence speed purposes. As we will show, the TITAN framework from \cite{Hien2023titan} allows building and analysing such an algorithm. Unfortunately, the theoretical convergence properties of TITAN assume exact proximal computations at each step, which cannot be ensured in practice in our context. To circumvent this, we thus propose and prove the convergence of an inexact version of a TITAN-based optimisation scheme. Inexact rules in the form of those studied in \cite{BCVMFB} are considered. We refer to the proposed method as to a Preconditioned Semi-Linearised Structure Adapted Proximal Alternating Minimisation (P--SASL--PAM) scheme.  {We investigate the convergence properties for this algorithm by relying on the K\L{} property first considered in \cite{attouch-bolte-svaiter}. Under analytical assumptions on the objective function, we show the global convergence toward a critical point of any sequence generated by the proposed method.} Then, we explicit the use of this method in our problem of image recovery and feature extraction. The performance of the approach is illustrated by means of examples in the field of image processing, in which we also show quantitative comparisons with state-of-the-art methods.\\

In a nutshell, the contributions of this work are (i) the proposition of an original non-convex variational model for the joint image recovery and feature extraction problem; (ii) the design of an inexact block coordinate descent algorithm to address the resulting minimisation problem; (iii) the convergence analysis of this scheme; (iv) {   the illustration of the performance of the proposed method through a numerical example in the field of ultrasound image processing.}\\

}

The paper is organised as follows. In \autoref{sec:model} we introduce the degradation model and report our derivation of the objective function for image recovery and feature extraction, starting from statistical assumptions on the data. In \autoref{sec:method}, we  describe the proposed P-SASL-PAM method to address a general non-smooth non-convex optimization problem; secondly we show that the proposed method converges globally, in the sense that the whole generated sequence converges to a (local) minimum. The application of the P-SASL-PAM method to the joint reconstruction/segmentation problem is described in \autoref{sec:application}. Some illustrative numerical results are shown in \autoref{sec:simulation}. Conclusions are drawn in \autoref{sec:conclusions}.\\

\section{Model Formulation}
\label{sec:model}
{In this section, we  
describe the construction of the objective function associated to the joint reconstruction/feature extraction problem.} After defining the degradation model, we report the Bayesian model that is reminiscent from the one considered in \cite{PPULA,US_GGD} in the context of ultrasound imaging. Then, we describe the procedure that leads us to the definition of our addressed optimization problem. 





\subsection{Observation Model}
Let $x\in\Rm^n$ and $y\in\Rm^m$ be respectively the vectorised sought-for {solution} and the observed data, which are assumed to be related according to the following model
\begin{equation}
    y = Kx + \omega,
    \label{eq:model}
\end{equation}
where $K \in \Rm^{m \times n}$ is a linear operator, and $\omega \sim \mathcal{N}(0, \sigma^2 \mathbb{I}_m)$, \emph{i.e.}\  the normal distribution with zero mean and covariance matrix $\sigma^2\mathbb{I}_m $ with $\sigma>0$ and $\mathbb{I}_m$ states for the $m\times m$ identity matrix.  We further assume that $x$ can be characterised by a finite set of $k$ \textit{features} that are defined in a suitable space, where the data are described by a simple model relying on a small number of parameters. 
The Generalised Gaussian Distribution ($\mathcal{GGD}$) 
\begin{multline}
(\forall t \in \mathbb{R})\quad 
\mathsf{p}(t;p, \alpha)\\= \frac{1}{2\alpha^{1/p}\Gamma \left(1 + \frac{1}{p}\right)}\exp\left(-\frac{|t|^p}{\alpha}\right)
\end{multline}
\label{eq:GGD}
{with $(p,\alpha)\in (0,+\infty)^2$} has shown to be a suitable and flexible tool for this purpose \cite{PPULA,  WAVELET_GGD,US_GGD}. 
Each feature can be identified by a pair $(p_j,\alpha_j)$ for $j\in \{1,\dots,k\}$, where  parameter $p$ is proportional to the decay rate of the tail of
the probability density function (PDF) and parameter $\alpha$ models the width of the peak
of the PFD. Taking into account the role that $p$ and $\alpha$ play in the definition of the PDF profile, these two parameters are generally referred to as \textit{shape} and \textit{scale} parameter.


Assuming that $K$ and $\sigma$ are known, the task we address in this work is to jointly retrieve $x$ (reconstruction) and obtain a good representation of its features through an estimation of the underlying model parameters $(p_j,\alpha_j)$ for $j\in \{1,\dots,k\}$ (feature extraction). 
Starting from a similar statistical model as the one considered in \cite{PPULA,US_GGD}, we infer a continuous variational framework which does not rely on the \textit{a priori} knowledge of the exact number of features $k$. We derive this model by performing a \textit{Maximum a Posteriori} estimation, which allows us to formulate the Joint Image Reconstruction and Feature Extraction task as a non-smooth and non-convex optimisation problem involving a coupling term and a block-coordinate separable one. \\

\subsection{Bayesian Model}

From \eqref{eq:model}, we derive the following likelihood
\begin{multline}
   \mathsf{p}(y|x,\sigma^2) \\=\frac{1}{(2\pi\sigma^2)^{n/2}} \exp\left( - \frac{\|y-Kx\|^2}{2\sigma^2}\right).
    \label{eq:prior:y}
\end{multline}
Assuming then that the components of $x$ are independent conditionally
to the knowledge of their feature class, 
{  the prior distribution of $x$ is a mixture of $\GGD$s}
\begin{multline}
   \mathsf{p}(x|p, \alpha)\\=\prod_{j=1}^k \frac{1}{\left(2\alpha^{1/p_j}_j\Gamma \left(1 + \frac{1}{p_j}\right)\right)^{N_j}}\exp\left(-\frac{\|\overline x_j\|^{p_j}_{p_j}}{\alpha_j}\right).
    \label{eq:prior:x}
\end{multline}
Hereabove, for every $x\in \Rm^n$ and a feature labels set $j\in\{1,\dots,k\}$, we define $\overline{x}_j\in \Rm^{N_j} $  as the vector containing only the $N_j$ components of $x$ that belong to the $j$-th {feature}. 
{Following the discussion in \cite{chaari2010hierarchical}, for the shape parameter, we choose a uniform distribution on a certain interval $[a,b]\subset [0,+\infty)$:

\begin{align}
    \mathsf{p}(p) & = \prod_{j=1}^k \frac{1}{b-a}\mathsf{I}_{[a,b]}(p_j).
    \label{eq:prior:shape}
    \end{align}
    This choice stems from the fact that setting $a=0$ and $b=3$ allows covering all possible values of the shape parameter encountered in practical applications, but no additional information about this parameter is available.
    For the scale parameter, we adopt 
 the Jeffreys distribution to reflect the lack of knowledge about this parameter:
    \begin{align}
     \mathsf{p}(\alpha) & = \prod_{j=1}^k \frac{1}{\alpha_j} \mathsf{I}_{[0,+\infty)}(\alpha_j). \label{eq:prior:scale}
\end{align}
 Note that such kind of prior is often used for scale parameters \cite{jeffreys1946invariant}.
   }
{Hereabove, $\mathsf{I}_{S}$ represents the characteristic function of some subset $S\subset \mathbb{R}$, which is equal to 1 over $S$, and 0 elsewhere.
}\\

\subsection{Variational Model}
In order to avoid to define \textit{a priori} the number of features, we regularise the problem by considering the 2D Total Variation (TV) of the $\GGD$ parameters $(p,\alpha)\in (0,+\infty)^n\times(0,+\infty)^n$. 
The idea of using Total Variation to define a segmentation procedure is studied in \cite{ SLAT, PC_MS_ROF,2stagsegm,CVX_APPROACH,2stagesegmblurry,Pascal2021AutomatedDS} by virtue of the co-area formula: the authors propose to replace the boundary information term of the Mumford-Shah (MS) functional \cite{Mumford1989OptimalAB} with the TV convex integral term. This choice yields a non-tight convexification of the MS model that does not require setting the number of segments in advance. The overall segmentation procedure is then built upon two steps: the first one consists of obtaining a smooth version of the given image that is adapted to segmentation by minimising the proposed functional with convex methods; the second step consists of partitioning the obtained solution into the desired number of segments, by \textit{e.g.} defining the thresholds with Otsu's method \cite{otsu} or the $k$-means algorithm. 
{The strength of our approach is that the second step (\textit{\emph{i.e.}\ } the actual segmentation step) is independent from the first one; hence it is possible to set the number of segments (\emph{i.e.}\ , labels) without solving the optimisation problem again.}

In the considered model, the introduction of a TV prior leads to a minimization problem that is non-convex with respect to $\alpha$. {To circumvent this issue, a possible strategy would involve applying the variable change $\beta = 1/\alpha$, which leads to a convex problem with regard to $\beta$. However, after performing some tests, we noticed that this choice tends to promote extreme values $0$ or $+\infty$. We then opted for the following reparameterisation for the scale parameter: let $\beta = (\beta_i)_{1 \leq i \leq n} \in \Rm^n$ be such that, for every $i \in \{1,\ldots,n\}$,
\begin{equation}
    \beta_i = \frac{1}{p_i}\ln \alpha_i,
\end{equation}
and let us choose for this new variable a non-informative Gaussian prior that is defined on the whole space of configurations with mean $\mu_{\beta}\geq 0$ and standard deviation $\sigma_{\beta}>0$. The choice of a non-necessarily zero-mean distribution stems from the idea of having a more flexible prior to represent our reparameterised scale parameter.}

Thus, replacing $\alpha$ with $\beta$ and further introducing TV regularisation potentials (weighted by the regularisation parameters $\lambda >0$ and $\zeta >0$) leads to the following reformulation of distributions \eqref{eq:prior:x}-\eqref{eq:prior:scale}:

\begin{multline}
    \mathsf{p}(x|p, \beta) \\= \prod_{i=1}^{n} \frac{1}{2\exp(\beta_i)\Gamma \left(1 + \frac{1}{p_i}\right)}\exp\left(-|x_i|^{p_i} \exp(-p_i \beta_i)\right)
    \label{eq:prior:x:2}
    \end{multline}
    
    
    
{\begin{equation}
       \mathsf{p}(p)  = c_p \exp(-\lambda \mathrm{TV}(p))\prod_{i=1}^n \frac{1}{b-a}\mathsf{I}_{[a,b]}(p_i)
    \label{eq:prior:shape:2}
    \end{equation}
    \begin{multline}
       {
   \mathsf{p}(\beta)} \\=  c_\beta  {\exp(-\zeta \mathrm{TV}(\beta ))\prod_{i=1}^n \frac{1}{\sqrt{2\pi}\sigma_{\beta}}\exp\Big(-\frac{(\beta_i-\mu_{\beta})^2}{2\sigma_{\beta}^2}\Big).}
    \label{eq:prior:shape:3}
    \end{multline}
    where $(c_p,c_\beta) \in (0,+\infty)^2$ are normalisation constants. 
In \autoref{fig:bayes}, we depict the probabilistic dependence graph defining the relations between variables and hyperparameters in our model.\\
\begin{figure}
\centering
\includegraphics[width=0.45\textwidth]{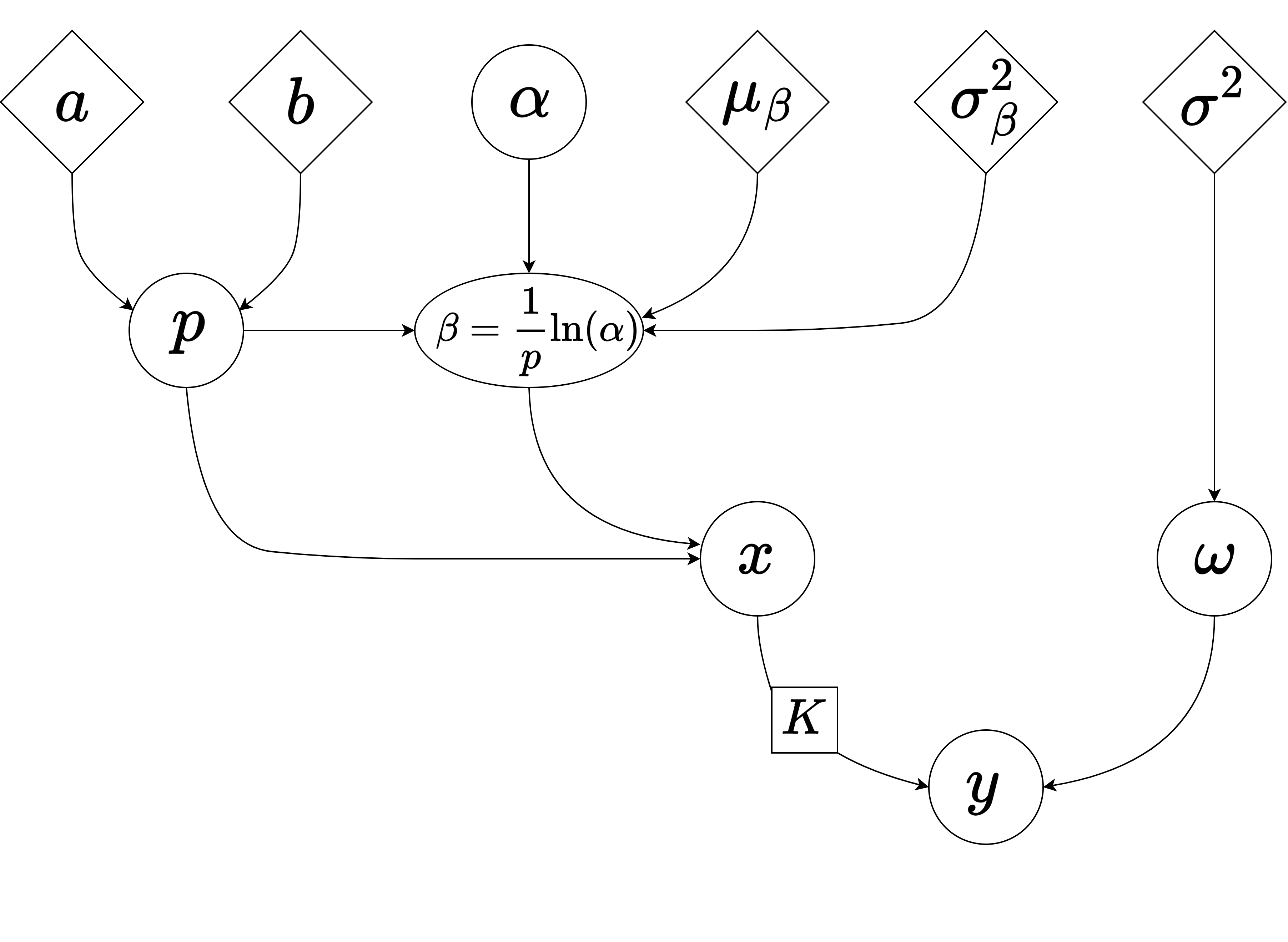}
\caption{Probabilistic dependence graph of our model. Hyperparameters are represented as diamonds, and variables as ellipses: $a$ and $b$ are the lower and the upper bound for the interval appearing in the uniform distribution of the shape parameter, $p$ is the shape parameter, $\alpha$ is the original scale parameter, $\beta$ is the reparameterised scale parameter with mean $\mu_{\beta}$ and standard deviation $\sigma_\beta$, $x$ is the sought signal,  $y$ is the observed one, $K$ is the linear operator, and $\omega$ is the additive Gaussian noise with standard deviation $\sigma$.}\label{fig:bayes}
\end{figure}}


The joint posterior distribution is determined as follows:
\begin{align}
\notag
   \mathsf{p}(x,p,\beta|y) &\propto\mathsf{p}(y|x,p,\beta)\mathsf{p}(x,p,\beta)\\
    & \propto\mathsf{p}(y|x,p,\beta)\mathsf{p}(x|p,\beta)\mathsf{p}(p)\mathsf{p}(\beta).    \label{eq:poster:distrib}
\end{align}
Let us take the negative logarithm of \eqref{eq:poster:distrib}, then 
computing the Maximum a Posteriori estimates (\emph{i.e.},
maximising the joint posterior distribution) is equivalent to the following optimization problem, which we refer to as the joint image reconstruction and feature extraction problem
\begin{multline}\label{eq:MAPNEGLOG}
     \minimize{(x,p,\beta) \in \Rm^n\times\Rm^n\times\Rm^n }{} \Theta(x,p,\beta) = \frac{1}{2\sigma^2} \|y-Kx\|^2 \\ + \sum_{i=1}^n  \Big( |x_i|^{p_i}e^{-p_i\beta_i} + \ln\Gamma(1 + \frac{1}{p_i}) + \iota_{[a,b]}(p_i) + \beta_i \\ +{ \frac{(\beta_i-\mu_{\beta})^2}{2\sigma_{\beta}^2}}\Big) + \lambda \mathrm{TV}(p) + \zeta \mathrm{TV}(\beta).
\end{multline}
{  Hereabove, $\iota_{S}$ represents the indicator function of some subset $S\subset \mathbb{R}$, which is equal to $0$ over $S$, and $+\infty$ elsewhere.\\
}

{ In \cite{SLPAM}, the authors proposed a generalised discrete Mumford-Shah variational model that is specifically designed for the joint image reconstruction and edge detection problem.
In contrast, the model we propose in \eqref{eq:MAPNEGLOG} is well suited to encompass a wider class of problems. In \autoref{sec:simulation}, we present two applications, namely in the context of wavelet-based image restoration and in the context of joint deblurring/segmentation of ultrasound images. In particular, we notice that when restricted to variable $x$ for a given set of parameters $(p,\beta)$, the formulation \eqref{eq:MAPNEGLOG} boils down to the flexible sparse regularisation model
\begin{equation}
    \minimize{x \in \Rm^n}{\frac{1}{2\sigma^2} \|y-Kx\|^2 + \sum_{i=1}^n  |x_i|^{p_i}e^{-p_i\beta_i}},
\label{eq:flexreg}
\end{equation}
where the contribution of the $\ell_{p_i}$ regularisation term is itself weighted in a 
space varying fashion.\\

}

{Function $\Theta$ in \eqref{eq:MAPNEGLOG} is non-smooth and non-convex. It reads as the sum of a coupling term and three block-separable terms. In particular, the block-separable data-fit term relative to $x$ is quadratic and hence has a Lipschitz continuous gradient. Our proposed algorithm aims to leverage this property, which is generally not accounted for by other BCD methods. To this aim, we exploit a hybrid scheme that involves both standard and linearised proximal steps. The details about the proposed method are presented in the next section.}\\

\section{Preconditioned Structure Adapted Semi-Linearised  Proximal Alternating Minimisation (P-SASL-PAM)}\label{sec:method}
In this section, we introduce a BCD-based method to address a class of sophisticated optimization problems that includes \eqref{eq:MAPNEGLOG} as a special case. We start the section by useful preliminaries about subdifferential calculus. Then, we present the \textit{Kurdyka-{\L}ojasiewicz property}, which plays a prominent role in the convergence analysis of BCD methods in a non-convex setting. Finally, we define problem \eqref{eq:objective}, itself generalising \eqref{eq:MAPNEGLOG}, for which we derive our proposed BCD-based algorithm and show its convergence properties. The so-called Preconditioned Structure Adapted Semi-Linearised  Proximal Alternating Minimisation (P-SASL-PAM) approach mixes both standard and preconditioned linearised proximal regularisation on the different coordinate blocks of the criterion.
\subsection{Subdifferential Calculus}
Let us now recall some definitions and elements of subdifferential calculus that will be useful in the upcoming sections. For a proper and lower semicontinuous function $h:  \mathbb{R}^n \rightarrow(-\infty,\infty]$, the domain of $h$ is defined as 
\begin{equation*}
    \operatorname{dom} h \;=\left\{u\in \mathbb{R}^n \mid h(u) < +\infty \right\}.
\end{equation*}
Firstly, we recall the notion of subgradients and subdifferential for convex functions.
\begin{definition}[Subgradient of a convex function] Let $h:  \mathbb{R}^n \rightarrow(-\infty,\infty]$ be a proper convex lower semicontinuous function. The subdifferential $\partial h(u^+)$ of $h$ at $u^+ \in \Rm^n$ is the set of all vectors $r\in\mathbb{R}^n$, called subgradients of $h$ at $u^+$, such that 
\begin{equation*}
    \forall u \in \mathbb{R}^n\;\; h(u) \geq h(u^+) + \langle r,u-u^+\rangle.
\end{equation*}
If $u^+\notin \operatorname{dom}h$, then $\partial h(u^+) = \varnothing$.
\end{definition}
Secondly, we consider the more general notion of (limiting)-subdifferential for non-necessarily convex functions, as proposed in {\cite[Definition 8.3]{rockafellar}}.
\begin{definition}[Limiting Subdifferential]
Let $h:\mathbb{R}^n \rightarrow(-\infty, +\infty]$ be a proper and lower semicontinuous function. For a vector $u^+\in \mathbb{R}^n$,
\begin{itemize}
    \item the Fréchet subdifferential of $h$ at $u^+$, written as $\hat{\partial}h(u^+)$, is the set of all vectors $r\in\mathbb{R}^n$ such that 
    \begin{equation*}
   h(u) \geq h(u^+) + \langle r,u-u^+\rangle + o(\|u-u^+\|);
    \end{equation*}
    if $u^+\notin \text{dom\;}h$, then $\hat\partial h(u^+) = \varnothing$;
    \item the limiting-subdifferential of $h$ at $u^+$, denoted by $\partial h (u^+)$, is defined as
    \begin{multline*}
  \partial h(u^+) = \{r \in \mathbb{R}^n\;|\; \exists\, u^k\rightarrow u^+,\\  h(u^k)\rightarrow h(u^+), r^k\rightarrow r, r^k \in \hat\partial h(u^k)\}.
    \end{multline*}
\end{itemize}
\end{definition}
If $h$ is lower semicontinuous and convex, then the three
previous notions of subdifferentiality are equivalent, \emph{i.e.}\  $\hat \partial h(u^+) = \partial h(u^+)$. If $h$ is differentiable, then $\partial h(u^+) = \{\nabla h(u^+)\}$. Now, it is possible to formalise the notion of critical points for a general function:
\begin{definition}[Critical point] Let $h:\mathbb{R}^n \rightarrow (-\infty,+\infty]$ be a proper function. A point $u^*\in \mathbb{R}^n$ is said to be a critical (or stationary) point for $h$ if $0\in\partial h(u^*)$.
\end{definition}
Eventually, we define the notion of proximal maps relative to the norm induced by a positive definite matrix. 
{
\begin{definition}
Let $\mathcal{S}_n$ be the set of symmetric and positive definite matrices in $\Rm^{n\times n }$. For a matrix $A\in \mathcal{S}_n$, the weighted $\ell_2$-norm induced by $A$ is defined as
\begin{equation}
(\forall u \in \Rm^n)\quad 
    \|u\|_A = (u^\top A u)^{1/2}.
\end{equation}
\end{definition}}
\begin{definition}
Let $h:\Rmn\rightarrow (-\infty,+\infty]$ be a proper and lower semicontinuous function, let $A\in \mathcal{S}_n$ and $u^+\in\mathbb{R}^n$. The proximity operator of function $h$ at $u^+$ with respect to the norm induced by $A^{}$ is defined as
\begin{equation}
    \prox_h^A (u^+) = \argmin _{u\in \Rmn} \left(\frac{1}{2} \|u-u^+\|^2_A + h(u)\right).
\end{equation}
\end{definition}
Note that $\prox_h^A (u^+)$, as defined above, can be the empty set. It is nonempty for
every $u^+\in \Rmn$, if $h$ is lower-bounded by an affine function. In addition, it reduces to a single-valued operator when $h$ is convex.

{  In order to deal with the situation when no closed-form proximal formulas are available (as it might be the case for non-trivial preconditioning metrics $A$), we take into account an inexact notion of proximal computation in the sense of \cite[Theorems 4.2 and 5.2]{attouch-bolte-svaiter} and \cite{BCVMFB}:

\begin{definition}
    Let $h:\Rmn\rightarrow (-\infty,+\infty]$ be a proper and lower semicontinuous function, let $A\in \mathcal{S}_n$, $\tau> 0$ and $u^+\in\mathbb{R}^n$. Then $u^*\in\R^n$ is an inexact proximal point for $h$ at $u^+$ if the following \emph{relative error} conditions are satisfied:
    \begin{enumerate}
        \item [(i)] \emph{Sufficient Decrease Condition:}
        \begin{equation}\label{eq:def:suff_dec}
            h(u^*) + \frac{1}{2}\|u^+ - u^*\|_{A}^2 \leq h(u^+)
        \end{equation}
        \item [(ii)] \emph{Inexact Optimality}: there exists $r \in \partial h(u^*)$ such that
        \begin{equation}\label{eq:def:inex_opt}
            \|r\| \leq \tau \|u^+ - u^*\|. 
        \end{equation}
    \end{enumerate}
    In this case we write that $u^* \in \operatorname{prox}_h^{A,\tau}(u^+)$.
\end{definition}
\begin{remark}
    We highlight that when exact proximal points are considered, the optimality condition reads as
    \begin{equation}
         0 \in \partial h(u^*) + A(u^+ - u^*)    \end{equation}
    implying that there exists $r \in \partial h(u^*)$ such that $r = A(u^*-u^+)$.
    
\end{remark}
}
\subsection{The K\L-Property}Most of the works related to BCD-based algorithms rely on the framework developed by Attouch, Bolte, and Svaiter in their seminal paper \cite{attouch-bolte-svaiter} in order to prove the convergence of block alternating strategies for non-smooth and non-convex problems. A fundamental assumption in \cite{attouch-bolte-svaiter} is that the objective function satisfies the \textit{Kurdyka-Łojasiewicz} (K{\L}) property \cite{KURDYKA,LOJA2, LOJA1}. We recall the definition of this property as it was given in \cite{PALM}. Let $\eta\in(0,+\infty]$ and denote by $\Phi_\eta$ the class of concave continuous functions ${\varphi : [0,+\infty) \rightarrow [0,+\infty)}$ satisfying the following conditions:
\begin{enumerate}[(i)]
    \item $\varphi(0) = 0$;
    \item $\varphi$ is $\mathcal{C}^1$ on $(0,\eta)$ and continuous at 0;
    \item for every $s\in (0,\eta)$, $\varphi'(s)>0$.
\end{enumerate}

 For any subset $S\subset \mathbb{R}^n$ and any point $u^+\in\mathbb{R}^n$, the distance from $u^+$ to $S$ is defined by  \begin{equation*}
     \operatorname{dist}(u^+,S) = \inf_{u\in S} \|u^+-u\|
 \end{equation*}
with $\operatorname{dist}(u^+,\varnothing)=+\infty$.
\begin{definition}[K{\L} property] Let $h:\Rmn\rightarrow (-\infty,+\infty]$ be a proper and lower semicontinuous function.
\begin{enumerate}[(i)]
    \item Function $h$ is said to satisfy the Kurdyka-Łojasiewicz property at ${u}^+\in \text{dom}\;\partial h$ if there exist $\eta\in (0,+\infty]$, a neighbourhood $U$ of ${u}^+$ and a function $\varphi \in \Phi_{\eta}$ such that, for every
    $u\in U$,
    \begin{multline}
    h(u^+)<h(u)<h(u^+) + \eta\\
    \quad \Rightarrow 
  \varphi'(h(u)-h(u^+))\operatorname{dist}(0,\partial h(u))\geq 1.
    \end{multline}
    \item Function $h$ is said to be a K{\L} function if it satisfies the K{\L} property at each point of $\operatorname{dom}\partial h$.
\end{enumerate}

\end{definition}

\subsection{Proposed Algorithm}
Let us consider that every element $\zeta \in \R^N$ is block-decomposed as $\zeta = (\zeta_0,\dots, \zeta_d)$, with, for every $i \in \{0,\ldots,d\}$, $\zeta_i \in\R^{n_i}$, with $\sum_{i=0}^d n_i = N$. As we will show in \autoref{sec:smoothing}, Problem \eqref{eq:MAPNEGLOG} is a special instance of the general class of problems of the form
\begin{equation}
    \label{eq:objective}
    \minimize{\zeta\in\R^N}{\theta(\zeta) = q(\zeta) + f(\zeta_0)  + \sum_{i=1}^d g_i(\zeta_i)},
\end{equation}
under the following assumption:
\begin{assumption}\label{assu:model3blocks}
$\phantom{-}$\\
\begin{enumerate}
\item Function $q:\R^N \to \R$ is bounded from below and differentiable with Lipschitz continuous gradient on bounded subsets of $\R^N$. 
\item Function $f: \R^{n_0} \to \R$ is differentiable with globally Lipschitz continuous gradient of constant $L_f>0$,
and is bounded from below. 
\item For every $i\in \{1,\dots,d\}$, function $g_i: \R^{n_i} \to (-\infty,+\infty]$ is proper, lower semicontinuous and bounded from below and the restriction to its domain is continuous. \label{assumption:model:gh}
\item {$\theta$ is a K{\L} function.\\} \label{assumption:model:KL}
\end{enumerate}
\end{assumption}

\begin{remark}
The assumption of continuity in \autoref{assu:model3blocks}.\ref{assumption:model:gh} is standard in the context of inexact minimisation algorithm (see the assumptions in {\cite[Theorem 4.1, Theorem 5.2]{attouch-bolte-svaiter}}).\\
\end{remark}

Throughout the paper we will use the following notation: for every $(\zeta_{i'})_{1\le i'\le d} \in \R^{n_1}\times \cdots \R^{n_d}$ and
$i\in \{0,\dots,d\}$, 
${\zeta_{\neq i} = (\zeta_0, \dots, \zeta_{i-1},\zeta_{i+1},\dots, \zeta_d)}$
and 
\begin{multline}
    (\forall z \in \R^{n_i})\quad (z;\zeta_{\neq i})\\= (\zeta_0, \dots, \zeta_{i-1}, z,\zeta_{i+1},\dots, \zeta_d).
\end{multline}
In order to proceed with the algorithm construction and analysis, let us recall the notion of partial subdifferentiation for a function $\fonc{\theta}{\R^N}{\R}$ as the one in \eqref{eq:objective}. For every $i\in \{0,\dots,d\}$ given a fixed $\zeta_{\neq i}$,
the subdifferential of the partial function $\theta(\cdot\,; \zeta_{\neq i})$ with respect to the $i$-th block, is denoted as $\partial_{i} \theta(\cdot\,; \zeta_{\neq i})$. Given these definitions, we have the following differential calculus property
(see \cite[Exercises 8.8(c), Proposition 10.5]{rockafellar}:\\

\begin{proposition}
Let function $\theta$ be defined as in \eqref{eq:objective}. Under \autoref{assu:model3blocks}, the following equality holds: for every $\zeta \in \mathbb{R}^N$,


\begin{multline}
 \partial\theta (\zeta)\\
 = \{\nabla_{0}q(\zeta) + \nabla f(\zeta_0)\} \times \cart_{\!i=1}^{\!d}\left(\nabla_{i}q(\zeta) + \partial g_i (\zeta_i)\right)\\= \cart_{\!i=0}^{\!d} \partial_{i} \theta(\zeta). 
\end{multline}
\label{prop:partial_subdifferential}
\end{proposition}

We are now ready to introduce our block alternating algorithm P-SASL-PAM, outlined in \autoref{alg:ours}, to solve problem \eqref{eq:objective}. Throughout the paper, we use the following notation: for every $\ell\in\N$ and and for $i\in \{1,\dots,d\}$,
\begin{equation*}
\begin{aligned}
\zeta^{\ell+1,0} &= \zeta^{\ell};\\ \zeta^{\ell+1,i} &= (\zeta^{\ell+1}_0, \dots,\zeta^{\ell+1}_{i-1}, \zeta^{\ell}_i, \zeta^{\ell}_{i+1}, \dots, \zeta^{\ell}_{d}  )\\ \zeta^{\ell+1,d+1} &= \zeta^{\ell+1}.
\end{aligned}
\end{equation*}

\begin{algorithm}[htb]
 \textbf{Initialize} $\zeta_0^0\in\dom f$, $\zeta_i^0\in\dom g_i$ for $i\in \{1,\dots,d\}$\;\\
 \textbf{Set} $(A_\ell)_{\ell \in \mathbb{N}} \in\mathcal{S}_{n_0}$ for every $\ell \in\N$\;\\
 \textbf{Set} $\gamma_0 \in(0,1)$ and $\gamma_i>0$
 and $\tau_i>0$ for $i\in \{1,\dots,d\}$ and $\tau_i>0$ for $i\in \{0,\dots,d\}$\;\\
 \textbf{For} $\,\ell = 0,1,\ldots$ \textbf{until} convergence
 {
 \begin{align}
  \zeta_0^{\ell +1} &\in \prox^{A_{\ell}, \tau_0}_{\gamma_0 q{(\cdot\,;\zeta_{\neq 0}^{\ell})}}\big(\zeta_0^{\ell}-\gamma_0{A_{\ell}^{-1}}\nabla f(\zeta_0^{\ell})\big)\label{eq:step1}
   \end{align}
   \phantom{\textbf{For}}\textbf{For} $\,i = 1,\ldots, d$
   \begin{align}
    \zeta_i^{{\ell}+1} &\in \prox^{\tau_i}_{\gamma_i\theta{(\cdot\, ;\,\zeta^{{\ell}+1, i}_{\neq i})}}(\zeta_i^{{\ell}})\label{eq:step2}
   \end{align}
   \phantom{\textbf{For}}\textbf{end}\\
 \textbf{end}}
 
 \caption{P-SASL-PAM to solve \eqref{eq:objective}.}
 \label{alg:ours}
\end{algorithm}

The proposed method sequentially updates one of the coordinate blocks $(\zeta_0,\dots,\zeta_d)$ involved in function $\theta$, through proximal and gradient steps. Our algorithm P-SASL-PAM, summarised in \autoref{alg:ours}, mixes both standard and linearised proximal regularisation on the coordinate blocks as in SLPAM \cite{SLPAM}, while inverting the splitting in order {to gain more efficient proximal computations} as in ASAP \cite{ASAP, Tan2019InertialAG}.
On the one hand, the lack of global Lipschitz-continuity of $\nabla q$ prevents us from adopting BCVMFB \cite{BCVMFB}. On the other hand, the lack of differentiability for the whole set of block-separable functions prevents us from adopting ASAP \cite{ASAP, Tan2019InertialAG}.
Our approach takes full advantage of the Lipschitz differentiability assumption on $f$ to perform a linearised step for the update of variable $\zeta_0$, while the remaining $\zeta_i$'s are updated sequentially, according to a standard proximal step.  In addition, in order to accelerate the convergence, a preconditioned version of the linearised step is used, which relies on the MM-based variable metric forward-backward strategy introduced in \cite{VMFB}. The latter relies on the following technical assumptions:\\

\begin{assumption}
\label{assumption:majorisation}
    We choose a sequence of SPD matrices $(A_{\ell})_{\ell\in \N}$ in such a way that there exists $(\underline \nu,\overline \nu)\in(0,+\infty)^2$ such that, for every $\ell\in \N$,
    \begin{equation}
    \underline \nu \mathbb{I}_{n_0} \preceq A_{\ell} \preceq \overline{\nu}\mathbb{I}_{n_0}.
    \label{eq:eigs}
\end{equation}
\end{assumption}

\begin{assumption}\label{assu:quad_maj}
     The quadratic function defined, for every $\zeta_0^+\in\R^{n_0}$ and every SPD matrix $A\in\mathcal{S}_n$ satisfying \autoref{assumption:majorisation}, as
    \begin{multline}
   (\forall \zeta_0\in \R^{n_0})\quad \phi(\zeta_0,\zeta_0^+)\\= f(\zeta_0^+) + \langle\zeta_0-\zeta_0^+,\nabla f(\zeta_0^+)\rangle +\frac{1}{2}\|\zeta_0^+-\zeta_0\|_A^2
\label{eq:quad:maj}
\end{multline}
is a majorant function of $f$ at $\zeta_0^+$, \emph{i.e.}\ 
\begin{equation}
  (\forall \zeta_0\in \R^{n_0})\quad  f(\zeta_0) \leq \phi(\zeta_0,\zeta_0^+).
  \label{eq:majorant:f}
\end{equation}
\end{assumption}

\begin{remark}
    {Since $f$ satisfies \autoref{assu:model3blocks}, the \emph{Descent Lemma} {\cite[Proposition A.24]{Bertsekas/99}} applies, yielding
\begin{multline*}
     (\forall (\zeta_0,\zeta_0^+)\in \R^{n_0}\times\R^{n_0})\quad f(\zeta_0)\\\leq f(\zeta_0^+) + \langle\zeta_0-\zeta_0^+,\nabla f(\zeta_0^+)\rangle +\frac{L_f}{2}\|\zeta_0^+-\zeta_0\|^2.
\end{multline*}
This guarantees that the preconditioning matrix $A=L_f\mathbb{I}_{n_0}$ satisfies \autoref{assumption:majorisation}, with $\underline{\nu} = \overline{\nu} = L_f$. Apart from this simple choice for matrix $A$, more sophisticated construction strategies have been studied in the literature \cite{VMFB, erdogan_fessler,hunter_langer}. Practical choices of metrics for Problem \eqref{eq:MAPNEGLOG} will be discussed in \autoref{sec:simulation}, which is dedicated to numerical experiments.\\
}
\end{remark}
{  
\begin{remark} Alternative approaches to deal with the lack of global Lipschitz continuity of $\nabla q$ could involve a backtracking strategy as in \cite{Salzo2017_VM} or adaptive step sizes based on an estimate of the local smoothness of the function as in \cite{malitsky2020adaptive,latafat2023adaptive}.\\
\end{remark}}

{  \begin{remark} \textbf{Esentially Cyclic update rule} Even though \autoref{alg:ours} relies on a sequential update rule for the blocks of coordinates $i\in\{0,\dots, d\}$, an extension to a quasi-cyclic rule with interval $\overline{d}\geq d$ is possible. In this case, at each iteration, the index $i\in\{0,\dots,d\}$ of the updated block is randomly chosen in such a way that each of the $d$ blocks is updated at least once every $\overline{d}$ steps. \\
    
\end{remark}}
P-SASL-PAM involves the computation of three proximal operators, at each iteration $\ell \in \N$. As we will show in \autoref{sec:titan}, if these operators are exactly computed, P-SASL-PAM fits within the general algorithmic framework TITAN~\cite{Hien2023titan}, and, as such, inherits its convergence properties. 
The links between the exact and the inexact form of \autoref{alg:ours} is discussed in \autoref{sec:inexact1}. The convergence of the inexact form of \autoref{alg:ours} is shown in \autoref{sec:convergence}.

\subsubsection{Links between P-SASL-PAM and TITAN}
\label{sec:titan}

Let us show that the exact version of \autoref{alg:ours} is a special instance of the TITAN algorithm from \cite{Hien2023titan}. 
The scheme of TITAN relies on an MM strategy that, at each iteration, for each block of coordinates, minimizes a block surrogate function, \emph{i.e.}\  a majorizing approximation for the restriction of the objective function to this block. Let us define formally the notion of block surrogate function in the case of Problem \eqref{eq:objective}.\\

\begin{definition}[Block surrogate function]
Consider a function $\fonc{h}{\R^N}{\R}$.
For every $i\in \{ 0,\dots,d\}$, function ${h_i\,:\,\R^{n_i}\times \R^N \to \R}$ is called a \emph{block surrogate function} of $h$ at block $i$ if $(\zeta_i,\xi) \mapsto h_i(\zeta_i;\xi)$ is continuous in $\xi$, lower-semicontinuous in $\zeta_i$ and the following conditions are satisfied:
\begin{enumerate}
\itemsep0em
    \item[(i)] $h_i(\xi_i;\xi)= h(\xi)$ for every $\xi  \in \R^N$
    \item[(ii)] $h_i(\zeta_i;\xi)\geq h(\zeta_i;\xi_{\neq i})$ for all $\zeta_i\in \R^{n_i}$ and ${\xi\in\R^N}$ 
\end{enumerate}
Function $h_i(\cdot;\xi)$ is said to be a block surrogate function of $h$ at block $i$ in $\xi$.
The \emph{block approximation error} for block $i$ at a point $(\zeta_i,\xi) \in \R^{n_i}\times \R^N$ is then defined for every $i\in \{0,\dots,d\}$ as
$${\mathsf{e}_i(\zeta_i;\xi) := h_i(\zeta_i;\xi) - h(\zeta_i; \xi_{\neq i})}.$$ 

\end{definition}

Let us now show that each of the steps of \autoref{alg:ours} is actually equivalent to minimising an objective function involving a block surrogate function of the differentiable terms in $\theta$ for block $i\in \{0,\dots,d\}$ at the current iterate. 

Solving \eqref{eq:step1} in \autoref{alg:ours} is equivalent to solving
\begin{equation}
 \argmind{\zeta_0\in \R^{n_0}}{ h_0(\zeta_0; \zeta^\ell)}
 \end{equation}
where 
\begin{multline*}
(\forall \zeta_0\in\mathbb{R}^{n_0})\quad  h_0(\zeta_0; \zeta^\ell)= \\ q(\zeta_0;\zeta^\ell_{\neq 0}) + f(\zeta_0^\ell) + \langle \nabla f(\zeta_0^\ell),\zeta_0 -  \zeta_0^\ell \rangle + \frac{1 }{2\gamma_1}\|\zeta_0^\ell - \zeta_0\|^2 _{{A_{\ell}}},
\end{multline*}
is a surrogate function of $(q(\cdot\,; \zeta^\ell_{\neq 0}) + f$ by virtue of \autoref{assu:quad_maj}. 
Notice that function $h_0$ is continuous
on the block $(\zeta_0;\zeta) \in \R^{n_0}\times \R^N$.

Solving \eqref{eq:step2} for a certain block index $i\in \{1,\dots,d\}$ in \autoref{alg:ours} is equivalent to solving
\begin{equation}
 \argmind{\zeta_i\in \R^{n_i}}{ h_i(\zeta_i; \zeta^{\ell+1,i})+ g_i(\zeta_i)}
 \end{equation}
where the function
\begin{multline*}
    (\forall \zeta_i\in \mathbb{R}^{n_i})\qquad h_i(\zeta_i;\zeta^{\ell+1,i})= \\ q(\zeta_i;\zeta^{\ell+1,i}_{\neq i}) + \frac{1}{2\gamma_i}\|\zeta_i-\zeta_i^\ell\|^2 
\end{multline*}
is a proximal surrogate of function $q(\cdot\; ,\zeta^{\ell+1}_{\neq i})$  
at its $i$-th block
in $\zeta^{\ell+1,i}$. Note that function $h_i$ is continuous on $\R^N$.

In a nutshell, \autoref{alg:ours} alternates between minimization of problems involving block surrogates for the differentiable terms of function $\theta$, and, as such, can be viewed as a special instance of TITAN \cite{Hien2023titan}. This allows us to state the following convergence result for a sequence generated by \autoref{alg:ours}.\\

\begin{theorem}
    Let Assumptions \ref{assu:model3blocks}-\ref{assu:quad_maj} be satisfied.
    Assume also that the sequence $(\zeta^{\ell})_{\ell\in \N}$ generated by \autoref{alg:ours} is bounded. Then,
    \begin{enumerate}[i)]
    \item\label{finitelentheorem2} $\sum_{\ell=0}^{+\infty}\|\zeta^{\ell+1} - \zeta^\ell\| < +\infty$;
    \item $(\zeta^{\ell})_{\ell\in \N}$ converges to a critical point $\zeta^*$ for function $\theta$ in \eqref{eq:objective}.
    \end{enumerate}
\end{theorem}
\begin{proof}
We start the proof by identifying the three block approximation errors for the block surrogate functions at an iteration $\ell \in \mathbb{N}$:
\begin{multline*}
    (\forall \zeta_0\in\mathbb{R}^{n_0})\quad  \mathsf{e}_0 (\zeta_0;\zeta^\ell)\\= f(\zeta_0^\ell) + \langle \nabla f(\zeta_0^\ell),\zeta_0 -  \zeta_0^\ell \rangle + \frac{1 }{2\gamma_0}\|\zeta_0^\ell - \zeta_0\|^2_{A_{\ell}} - f(\zeta_0)
    \end{multline*}
    and for $i\in \{1,\dots,d\}$
    \begin{multline*}
     (\forall \zeta_i\in \mathbb{R}^{n_i})\quad \mathsf{e}_i(\zeta_i;\zeta^{\ell+1,i}) = \frac{1}{2\gamma_i}\|\zeta_i-\zeta_i^\ell\|^2.
\end{multline*}
Clearly $\mathsf{e}_0 (\cdot\,;\zeta^\ell)$ (resp.\ $\mathsf{e}_i(\cdot\,;\zeta^{\ell+1,i})$ for $i\in \{1,\dots,d\}$) are differentiable at $\zeta_0^\ell$ (resp. at $\zeta_i^\ell$ for $i\in \{1,\dots,d\}$) 
and the following holds for every $i\in \{0,\dots,d\}$:
\begin{equation*}
\begin{aligned}
 & \mathsf{e}_i(\zeta_i^\ell;\zeta^{\ell+1,i}) = 0,  && \nabla_{i}\mathsf{e}_i(\zeta_i^\ell;\zeta^{\ell+1,i}) = 0. \\
 \end{aligned}
 \end{equation*}
 This shows that {\cite[Assumption 2]{Hien2023titan}} is satisfied.

From \eqref{eq:step1} and Assumptions~\ref{assumption:majorisation}-\ref{assu:quad_maj}, we deduce that
\begin{multline*}
q(\zeta^{\ell+1,1})   + f(\zeta_0^{\ell+1}) + \frac{1}{2}\left(\frac{1}{\gamma_0} - 1\right)\underline{\nu}\|\zeta_0^{\ell+1}-\zeta_0^\ell\|^2 \\ \leq q(\zeta^{\ell}) +f(\zeta_1^\ell)
\end{multline*}
which implies that the Nearly Sufficient Descending Property {\cite[(NSDP)]{Hien2023titan}} is satisfied for the first block of coordinate with constant $\frac{1}{2}\left(\frac{1}{\gamma_0} - 1\right)\underline{\nu}$. On the other hand, for every $i\in \{1,\dots,d\}$, function $\mathsf{e}_i(\cdot\,; \zeta^{\ell+1,i})$ satisfies {\cite[Condition 2]{Hien2023titan}}, which implies that {\cite[(NSDP)]{Hien2023titan}} also holds for $i$-the block of coordinates with the corresponding constant $1/\gamma_i$.

Moreover, \cite[Condition 4.(ii)]{Hien2023titan} is satisfied by \autoref{alg:ours} with $$\overline{l} = \max \left\{\frac{\underline{\nu}}{2}\left(\frac{1}{\gamma_0}-1\right), \frac{1}{\gamma_1},\dots, \frac{1}{\gamma_d} \right\}$$ and this constant fulfill the requirements of \cite[Theorem 8]{Hien2023titan}. In addition, by virtue of  Proposition~\ref{prop:partial_subdifferential}, \cite[Assumption 3.(i)]{Hien2023titan} holds, while the requirement in \cite[Assumption 3.(ii)]{Hien2023titan} is guaranteed by the fact that all the block surrogate functions are continuously differentiable.

Finally, since \autoref{alg:ours} does not include any extrapolation step, we do not need to verify \cite[Condition 1]{Hien2023titan}, whereas \cite[Condition 4.(i)]{Hien2023titan} is always satisfied.

In conclusion, we proved that all the requirements of  \cite[Proposition 5, Theorems 6 and 8]{Hien2023titan} are satisfied. \cite[Proposition 5]{Hien2023titan}  guarantees that the sequence has the finite-length 
property as expressed by \ref{finitelentheorem2}),
 while \cite[Theorems 6 and 8]{Hien2023titan} state that the sequence converges to a critical point $\zeta^*$ of \eqref{eq:objective}, which concludes the proof.
\end{proof}
\subsubsection{Well-definition of \autoref{alg:ours}} 
\label{sec:inexact1}

Now, we show that the inexact updates involved in \autoref{alg:ours} are well-defined. To do so, we prove that the P-SASL-PAM algorithm with exact proximal computations can be recovered as a special case of \autoref{alg:ours}. 

By the variational definition of proximal operator, for every $\ell \in \N$, the iterates of \autoref{alg:ours} satisfy, for every $i\in \{1,\dots,d\}$,
\begin{multline}
   \zeta_0^ {\ell+1} \in \argmind{u_0\in \R^{n_0}}{\Big\{q(u_0\, ;\zeta_{\neq 0}^\ell)+ \langle \nabla f(\zeta_0^\ell), u_0 - \zeta_0^\ell\rangle \\+ \frac{1}{2\gamma_0}\|u_0-\zeta_0^\ell\|_{A_{\ell}}^2 \Big\}}
   \end{multline}
   \begin{multline}
   \zeta_i^ {\ell+1} \in \argmind{u_i\in \R^{n_i}}{\Big\{q(u_i\, ;\zeta_{\neq i}^{\ell+1,i}) + g_i(u_i) \\+ \frac{1}{2\gamma_i}\|u_i-\zeta_i^\ell\|^2  \Big\}} 
\end{multline}
so that
\begin{multline}
q(\zeta^{\ell+1,1})  + \langle \nabla f(\zeta_0^\ell), \zeta_0^{\ell+1} - \zeta_0^\ell\rangle\\ + \frac{1}{2\gamma_0}\|\zeta_0^{\ell+1}-\zeta_0^\ell\|_{A_{\ell}}^2 \leq  q(\zeta^{\ell})
\end{multline}
\begin{multline}
     q(\zeta^{\ell+1,i+1}) + g(\zeta_i^{\ell+1})\\ + \frac{1}{2\gamma_i}\|\zeta_i^{\ell+1}-\zeta_i^\ell\|^2 \leq q(\zeta^{\ell+1,i}) + g(\zeta_i^{\ell}),  
\end{multline}
which implies that the sufficient decrease condition \eqref{eq:def:suff_dec} is satisfied for every $i\in \{0,\dots,d\}$.\\

The use of the Fermat's rule implies that, for every $\ell \in \N$, the iterates of P-SASL-PAM are such that for every $i\in \{1,\dots,d\}$ there exists $r_i \in \partial g_i(\zeta_i^{\ell+1})$ for which the following equalities are satisfied for $i\in \{1,\dots,d\}$:
\begin{align}
    & 0 = \nabla f(\zeta_0^{\ell}) +  \nabla_{0} q(\zeta^{\ell+1,1}) + \gamma_0^{-1}{A_{\ell}}(\zeta_0^{\ell+1}-\zeta_0^\ell) \\
    & 0 = r_i + \nabla_{i} q(\zeta^{\ell+1, i+1}) + \gamma_i^{-1}(\zeta_i^{\ell+1} - \zeta_i^{\ell})
\end{align}
Hence, for every $i\in \{1,\dots,d\}$
\begin{align}
     \|\nabla f(\zeta_0^{\ell}) +  \nabla_{0} q(\zeta^{\ell+1,1}) \| &\leq  \gamma_0^{-1}\overline{\nu}\|\zeta_0^{\ell+1}-\zeta_0^\ell\| \\
    \| r_i + \nabla_{i} q(\zeta^{\ell+1, i+1}) \| &= \gamma_i^{-1}\|\zeta_i^{\ell+1} - \zeta_i^{\ell}\|
\end{align}
which implies that the inexact optimality condition \eqref{eq:def:inex_opt} is satisfied with $\tau_0 = \gamma_0^{-1}\overline{\nu}$ for the first block of coordinates and  $\tau_i = \gamma_i^{-1}$ for the remaining ones. In a nutshell, \autoref{alg:ours} is well defined, as its inexact rules hold assuming exact computation of the proximity operators, which leads to TITAN.

\subsection{Convergence analysis in the inexact case}
\label{sec:convergence}

Let us now present our main result, that is the convergence analysis for \autoref{alg:ours}.\\


\begin{lemma}
\label{lem:suff_dec_iSASLPAM}
Let $(\zeta^\ell)_{\ell\in \N}$ be the sequence generated by \autoref{alg:ours}. Then, under  
Assumptions \ref{assu:model3blocks} and \ref{assumption:majorisation}
 \begin{enumerate}[i)]
 \item there exists $\mu\in(0,+\infty)$ such that for every $\ell \in \mathbb{N}$,
 \begin{equation}
 \label{eq:suff:dec}
     \theta(\zeta^{\ell+1}) \leq \theta(\zeta^{\ell}) - \frac{\mu}{2}\|\zeta^{\ell+1}-\zeta^{\ell}\|^2.
 \end{equation}
 \item $\sum_{\ell=0}^{+\infty}\|\zeta^{\ell+1}-\zeta^{\ell}\|^2 < +\infty$.
 \end{enumerate}
\end{lemma}
\begin{proof}
Let us start by considering the \emph{sufficient decrease inequality} related to the first block:
\begin{multline}\label{eq:suff_dec:x}
    q(\zeta^{\ell+1,1})  + \langle \nabla f(\zeta_0^\ell), \zeta_0^{\ell+1} - \zeta_0^\ell\rangle \\+ \frac{1}{2\gamma_0}\|\zeta_0^{\ell+1}-\zeta_0^\ell\|_{A_{\ell}}^2 \leq  q(\zeta^{\ell}).
\end{multline}
Adding $f(\zeta_0^\ell) + \frac{1}{2}\|\zeta_0^{\ell+1} - \zeta_0^{\ell}\|_{A_{\ell}}^2$ on both sides of \eqref{eq:suff_dec:x} yields
\begin{multline}\label{eq:suff_dec_inex:eq1}
    q(\zeta^{\ell+1,1})  + \langle \nabla f(\zeta_0^\ell), \zeta_0^{\ell+1} - \zeta_0^\ell\rangle \\+ \frac{1}{2\gamma_0}\|\zeta_0^{\ell+1}-\zeta_0^\ell\|_{A_{\ell}}^2 +f(\zeta_0^\ell) + \frac{1}{2}\|\zeta_0^{\ell+1} - \zeta_0^{\ell}\|_{A_{\ell}}^2 \\
    \leq  q(\zeta^{\ell}) + f(\zeta_0^\ell) + \frac{1}{2}\|\zeta_0^{\ell+1} - \zeta_0^{\ell}\|_{A_{\ell}}^2
\end{multline}

By applying \eqref{eq:quad:maj} and \eqref{eq:majorant:f} with $\zeta_0^+ = \zeta_0^\ell$ and $\zeta_0=\zeta_0^{\ell+1}$ we obtain 
\begin{multline}
    f(\zeta_0^{\ell+1}) \leq f(\zeta_0^{\ell}) + \langle \zeta_0^{\ell+1}-\zeta_0^\ell , \nabla f(\zeta_0^\ell)\rangle \\ +\frac{1}{2}\|\zeta_0^{\ell+1}-\zeta_0^\ell\|_{A_{\ell}}^2
\end{multline}
hence the LHS in \eqref{eq:suff_dec_inex:eq1} can be further lower bounded, yielding
\begin{multline}
    \label{eq:suff_dec_inex:eq2}
 q(\zeta^{\ell+1,1})   + f(\zeta_0^{\ell+1}) + \frac{1}{2\gamma_0}\|\zeta_0^{\ell+1}-\zeta_0^\ell\|_{A_{\ell}}^2 \\\leq q(\zeta^{\ell}) +f(\zeta_0^\ell)  + \frac{1}{2}\|\zeta_0^{\ell+1}-\zeta_0^\ell\|_{A_{\ell}}^2,
\end{multline}
hence 
\begin{multline}
    \label{eq:suff_dec_inex:eq3}
 q(\zeta^{\ell+1,1})   + f(\zeta_0^{\ell+1}) + \frac{1}{2}\left(\frac{1}{\gamma_0} - 1\right)\|\zeta_0^{\ell+1}-\zeta_0^\ell\|_{A_{\ell}}^2\\ \leq q(\zeta^{\ell}) +f(\zeta_0^\ell).
\end{multline}
To conclude, by \autoref{assumption:majorisation}, we get
\begin{multline}
    \label{eq:suff_dec_inex:eq4}
  q(\zeta^{\ell+1,1})   + f(\zeta_0^{\ell+1}) + \frac{1}{2}\left(\frac{1}{\gamma_0} - 1\right)\underline \nu\|\zeta_0^{\ell+1}-\zeta_0^\ell\|^2 \\ \leq q(\zeta^{\ell}) +f(\zeta_0^\ell).
\end{multline}
The sufficient decrease inequality for the remaining blocks of index $i\in \{1,\ldots,d\}$ can be expressed as  
\begin{multline}\label{eq:suff_dec:blocks}
    q(\zeta^{\ell+1,i+1}) + g(\zeta_i^{\ell+1}) - g(\zeta_i^{\ell})  + \frac{1}{2\gamma_i}\|\zeta_i^{\ell+1}-\zeta_i^\ell\|^2 \\\leq q(\zeta^{\ell+1,i}).
\end{multline}
The first term in the LHS of \eqref{eq:suff_dec:blocks} for the $i$-th block can be similarly bounded from below with the sufficient decrease inequality for the $(i+1)$-th block, yielding
\begin{multline}\label{eq:suff_dec:blocks:1.5}
    q(\zeta^{\ell+1,i+2}) + g(\zeta_{i+1}^{\ell+1})- g(\zeta_{i+1}^{\ell}) \\ + \frac{1}{2\gamma_{i+1}}\|\zeta_{i+1}^{\ell+1}-\zeta_{i+1}^\ell\|^2 + g(\zeta_i^{\ell+1}) - g(\zeta_i^{\ell}) \\ + \frac{1}{2\gamma_i}\|\zeta_i^{\ell+1}-\zeta_i^\ell\|^2 \leq q(\zeta^{\ell+1,i}).
\end{multline}
By applying this reasoning recursively from $i=1$ to $i=d$, we obtain
\begin{multline}\label{eq:suff_dec:blocks:2}
    q(\zeta^{\ell+1,d+1}) + \sum_{i=1}^dg(\zeta_i^{\ell+1}) - \sum_{i=1}^dg(\zeta_i^{\ell})  \\ + \sum_{i=1}^d\frac{1}{2\gamma_i}\|\zeta_i^{\ell+1}-\zeta_i^\ell\|^2 \leq q(\zeta^{\ell+1,1}) 
\end{multline}
where we recall that $q(\zeta^{\ell+1,d+1}) = q(\zeta^{\ell+1})$.

Exploiting now \eqref{eq:suff_dec:blocks:2}, we can lower bound the first term in the LHS of \eqref{eq:suff_dec_inex:eq4}, which yields
\begin{multline}
    \label{eq:suff_dec_inex:eq5}
  q(\zeta^{\ell+1}) + \sum_{i=1}^dg(\zeta_i^{\ell+1}) - \sum_{i=1}^dg(\zeta_i^{\ell}) \\ + \sum_{i=1}^d\frac{1}{2\gamma_i}\|\zeta_i^{\ell+1}-\zeta_i^\ell\|^2 + f(\zeta_0^{\ell+1}) \\ + \frac{1}{2}\left(\frac{1}{\gamma_0} - 1\right)\underline \nu\|\zeta_0^{\ell+1}-\zeta_0^\ell\|^2\leq q(\zeta^{\ell}) +f(\zeta_0^\ell).
\end{multline}

By setting $\mu = \min\left\{ \left(\frac{1}{\gamma_0} - 1\right)\underline{\nu}, \frac{1}{\gamma_1}, \dots, \frac{1}{\gamma_d}\right\}$, we deduce
\eqref{eq:suff:dec}.\\


From \eqref{eq:suff:dec}, it follows that the sequence $(\theta(\zeta^{\ell}))_{\ell\in \mathbb{N}}$ is non-increasing. Since function $\theta$ is assumed to be bounded from below, this sequence converges to some real number $\underline\theta$. We have then, for every integer $K$, \begin{equation}\begin{aligned}
    \sum_{\kappa=0}^K \|\zeta^\ell - \zeta^{\ell+1}\|^2 &\leq \frac{1}{\mu}\sum_{\kappa=0}^K \left(\theta(\zeta^\ell) - \theta(\zeta^{\ell+1}) \right) \\
    &= \frac{1}{\mu} (\theta(\zeta^0) - \theta(\zeta^{K+1}))\\
    & \le \frac{1}{\mu} (\theta(\zeta^0) - \underline\theta).
\end{aligned}\end{equation}
Taking the limit as $K\rightarrow +\infty$ yields the desired summability property.
\end{proof}

\begin{lemma}
Assume that the sequence  $(\zeta^{\ell})_{\ell \in \mathbb{N}}$  generated by \autoref{alg:ours} is bounded. Then, for every $\ell \in \mathbb{N}$,
 there exists $s^{\ell+1} \in \partial \theta(\zeta^{\ell+1})$ such that 
 \begin{equation}\label{eq:inexopt}
     \|s^{\ell+1}\| \leq \rho \|\zeta^{\ell+1} - \zeta^{\ell}\|,
 \end{equation}
 where $\rho\in (0,+\infty)$.
\end{lemma}
\begin{proof}
The assumed boundedness implies that there exists a bounded subset $S$ of $\R^N$ such that for every $i\in\{0,\dots,d\}$ and $\ell\in \mathbb{N}$,
 $\zeta^{\ell+1,i}\in S$.
For every $\ell \in \N$, we define
\begin{equation}
    s_0^{\ell+1} = \nabla f(\zeta_0^{\ell+1}) + \nabla_{0} q(\zeta^{\ell+1}) 
\end{equation}
for which the following holds by virtue of Proposition~\ref{prop:partial_subdifferential}
\begin{equation}
    s^{\ell+1}_0\in \partial_{0} \theta (\zeta^{\ell+1}) = \{\nabla_{0}\theta(\zeta^{\ell+1})\}.
\end{equation}

Then {\allowdisplaybreaks
\begin{align*}
    \|s^{\ell+1}_0\| 
     &\leq   \| \nabla f(\zeta_0^{\ell+1})  - \nabla f(\zeta_0^\ell)  \| \\&+ \|\nabla f(\zeta_0^\ell)  + \nabla_{0}  q(\zeta^{\ell+1,1}) \| \\&+ \|\nabla_{0} q(\zeta^{\ell+1})- \nabla_{0} q(\zeta^{\ell+1,1})\|.
\end{align*}}
 From the Lipschitz continuity of $\nabla f$ and $\nabla q$ on $S$ and the inexact optimality inequality for the first block, we conclude that
\begin{equation}
\|s^{\ell+1}_0\| \leq \left( L_f + \tau_0 + L_q\right)\|\zeta^{\ell+1} - \zeta^\ell\|.
\end{equation}
 
In the same spirit, for every $i\in \{1,\dots,d\}$ we consider $r_i^{\ell+1}\in \partial g(\zeta_i^{\ell+1})$ satisfying the inexact optimality inequality with the corresponding $\tau_i$. We then define
\begin{align}
    s_i^{\ell+1} &=  \nabla_{i} q(\zeta^{\ell+1}) + r_i^{\ell+1}  ÷\nonumber\\ &\in \nabla_{i} q(\zeta^{\ell+1}) + \partial g_i(\zeta_i^{\ell+1}) = \partial_{i} \theta(\zeta^{\ell+1}).
\end{align}
For $i=d$, by virtue of the inexact optimality inequality,
\begin{equation}
   \| s_d^{\ell+1}\| \leq \tau_d \|\zeta^{\ell+1} - \zeta^\ell\|.
\end{equation}
On the other side, for $i=1,\dots,d-1$
{\allowdisplaybreaks\begin{align*}
    \|s_i^{\ell+1}\| &= \| \nabla_{i} q(\zeta^{\ell+1}) + r_i^{\ell+1}\| \\ 
     &\leq  \|  \nabla_{i} q(\zeta^{\ell+1})- \nabla_{i} q(\zeta^{\ell+1,i+1})\| \\&\qquad \qquad + \| r_i^{\ell+1}  + \nabla_{i} q(\zeta^{\ell+1,i+1})\|\\
    &\leq  L_q\|\zeta^{\ell+1}-\zeta^\ell\| + \tau_i\|\zeta_i^{\ell+1}-\zeta_i^\ell\|,
\end{align*}}
where the last estimate stems from inexact optimality inequality for the $i$-th block. This yields
\begin{equation}
    \|s_i^{\ell+1}\| \leq (L_q + \tau_i)\|\zeta^{\ell+1} - \zeta^\ell\|.
\end{equation}
To conclude, setting
\begin{equation}
    s^{\ell+1} = (s_0^{\ell+1},\dots, s_d^{\ell+1})   \in \partial \theta(\zeta^{\ell+1})
\end{equation}
and $\rho = L_f + \sum_{i=0}^d \tau_i + dL_q$  yields \eqref{eq:inexopt}.
\end{proof}

We now report a first convergence result for a sequence generated by the proposed algorithm, which is reminiscent from \cite[Proposition 6]{PAM}:\\

 \begin{proposition}[Properties of the cluster points set]
 Suppose that Assumptions~\ref{assu:model3blocks} and  \ref{assumption:majorisation} hold. Let $(\zeta^\ell)_{\ell\in\mathbb{N}}$ be a sequence generated by \autoref{alg:ours}. Denote by $\omega(\zeta^0)$ the (possibly empty) set of its cluster points. Then
 \begin{enumerate}[i)]
     \item if $(\zeta^\ell)_{\ell\in\mathbb{N}}$ is bounded, then $\omega(\zeta^0)$ is a nonempty compact connected set and $$ \operatorname{dist}(\zeta^\ell,\omega(\zeta^0)) \rightarrow 0 \quad \text{\;as\;}\quad \ell\rightarrow +\infty;  $$
     \item $\omega(\zeta^0) \subset \operatorname{crit}\, \theta$, where $\operatorname{crit}\, \theta$ is the set of critical points of function $\theta$;
     \item $\theta$ is finite valued and constant on $\omega(\zeta^0)$, and it is equal to \[\inf_{\ell\in \mathbb{N}}\theta(\zeta^\ell) = \lim_{\ell \rightarrow +\infty} \theta(\zeta^\ell).\]
 \end{enumerate}
 \end{proposition}
 \begin{proof}
 The proof of the above results for the proposed algorithm is basically identical to the one for \cite[Proposition 6]{PAM} for PAM algorithm. In addition, we highlight that according to \autoref{assu:model3blocks}, our objective function $\theta$ is continuous on its domain.
 \end{proof}
 In conclusion, we have proved that, under Assumptions~\ref{assu:model3blocks}-\ref{assu:quad_maj}, a bounded sequence generated by the proposed method satisfies the assumptions in  \cite[Theorem 2.9]{attouch-bolte-svaiter}. Consequently, we can state the following result:\\

 \begin{theorem}\label{prop:main}
 Let Assumptions~\ref{assu:model3blocks}-\ref{assu:quad_maj} be satisfied and let $(\zeta^{\ell})_{\ell \in \mathbb{N}}  $ be a sequence generated by \autoref{alg:ours} that is assumed to be bounded. Then,
 \begin{enumerate}[i)]
     \item $\sum_{\ell=1}^{+\infty} \|\zeta^{\ell+1} - \zeta^\ell\|< +\infty$;
     \item  $(\zeta^{\ell})_{\ell \in \mathbb{N}} $ converges to a critical point $\zeta^*$ of $\theta$. \\
 \end{enumerate} 
 \end{theorem}

 We managed to show that both the exact and the inexact version of \autoref{alg:ours} share the same convergence guarantees under Assumptions~\ref{assu:model3blocks}-\ref{assu:quad_maj}. One of the main differences between the two algorithms, as highlighted in \cite{attouch-bolte-svaiter}, is that the former has convergence guarantees that hold for an objective function that is lower semicontinuous, whereas the latter requires its continuity on the domain. 
However, as it will be shown in the next section, this does not represent an obstacle to the use of \autoref{alg:ours} in image processing applications.\\
 
\color{black}
\section{Application of P-SASL-PAM
}\label{sec:application}

\subsection{Smoothing of the coupling term}\label{sec:smoothing}
 The application of \autoref{alg:ours} to Problem \eqref{eq:MAPNEGLOG} requires the involved functions to fulfil the requirements listed in \autoref{assu:model3blocks}.  This section is devoted to this analysis, by first defining $d=2$, $n_0=n_1=n_2 = n$, $N = 3n$ and the following functions,
 for every $x=(x_i)_{1\le i \le n}\in \Rmn$,
 $p = (p_i)_{1\le i \le n}\in \Rmn$, and $\beta = (\beta_i)_{1\le i \le n}\in \Rmn$,
 {\allowdisplaybreaks\begin{align}
 \tilde q(x,p,\beta) &= \sum_{i=1}^{n} |x_i|^{p_i}e^{-\beta_ip_i}, \label{eq:coupling}\\
  f(x) &= \frac{1}{2\sigma^2} \|y-Kx\|_2^2, \label{eq:f}
  \\
g_1(p) &= \sum_{i=1}^n\left(  \ln \Gamma(1+\frac{1}{p_i}) +
    \iota_{[a,b]}(p_i) \right) \label{eq:g}\\
    &\qquad \qquad +\lambda \operatorname{TV}(p),\notag \\
g_2(\beta) &= \sum_{i=1}^n\left(\beta_i+ {  \frac{(\beta_i - \mu_{\beta})^2}{2\sigma_\beta^2}} \right)  + \zeta \operatorname{TV}(\beta).\label{eq:h}
\end{align}}
The first item in \autoref{assu:model3blocks} regarding the regularity of the coupling term is not satisfied by \eqref{eq:coupling}. To circumvent this difficulty, we introduce the \textit{pseudo-Huber loss function} \cite{Charbonnier1997DeterministicER} depending on a pair of parameters {$\delta = (\delta_1,\delta_2) \in (0,+\infty)^2$ such that $\delta_2 < \delta_1$:
\begin{equation}
(\forall t \in \Rm)\quad
    C_{\delta}(t) = H_{\delta_1}(t)-\delta_2,
    \label{eq:Huber}
\end{equation}
where $H_{\delta_1}$ is the \textit{hyperbolic function} defined, for every $t \in \mathbb{R}$, by $H_{\delta_1}(t)=\sqrt{t^2 + \delta^2_1}$.
Function \eqref{eq:Huber} is used 
 as a smooth approximation of the absolute value involved in \eqref{eq:coupling}. 
 }
 We then replace \eqref{eq:coupling} with
\begin{equation}
    q(x,p,\beta) = \sum_{i=1}^{n} \left(C_\delta(x_i)\right)^{p_i}e^{-\beta_ip_i}. \label{eq:q}
\end{equation}
{Function $C_{\delta}$ is infinitely differentiable.
Thus function \eqref{eq:q} satisfies \autoref{assu:model3blocks}}. 

{
Function \eqref{eq:f} is quadratic convex, hence it clearly satisfies \autoref{assu:model3blocks}(ii). 
Function \eqref{eq:g} is a sum of functions that are {proper, lower semicontinuous} and either non-negative or bounded from below. The same applies to function \eqref{eq:h}, which is also strongly convex. It results that \eqref{eq:g} and \eqref{eq:h} satisfy \autoref{assu:model3blocks}(iii).\\}

Now, we must show that $\Theta$ {is a K\L{} function}. To do so, let us consider the notion of \textit{o-minimal structure} \cite{dries_1998}, which is a particular family $\mathcal{O}=\{\mathcal{O}_n\}_{n\in \mathbb{N}}$
where each $\mathcal{O}_n$ is a collection of subsets of $\Rmn$, satisfying a series of axioms (we refer to {\cite[Definition 13]{PAM}}, for a more complete description). We present hereafter the definition of \textit{definable set} and \textit{definable function} in an o-minimal structure:\\

\begin{definition}[Definable sets and definable functions]
Given an o-minimal structure $\mathcal{O}$, a set $\mathcal{A}\subset \Rmn$ such that $ \mathcal{A}\in \mathcal{O}_n$ is said to be definable in $\mathcal{O}$. A real extended valued function $f\,:\Rm \rightarrow (-\infty,+\infty]$ is said to be definable in $\mathcal{O}$ if its graph is a definable subset of $\Rmn\times\Rm$.\\
\end{definition}

The importance of these concepts in mathematical optimisation is related to the following key result concerning the K\L{} property \cite{ominimalstruct}:\\

\begin{theorem}
Any proper lower semicontinuous function $f: \mathbb{R}^n \rightarrow (-\infty,+\infty]$ which is definable
in an o-minimal structure $\mathcal{O}$ has the K\L{} property at each point of $\operatorname{dom}\,\partial f$.\\
\end{theorem}

Let us identify a structure in which all the functions involved in the definition of $\Theta$ are definable. This will be sufficient, as definability is a closed property with respect to several operations, including finite sum and composition of functions. Before that, we provide a couple of examples of o-minimal structure. The first is represented by the structure of \textit{globally subanalytic sets} $\Rm_{\rm an}$ \cite{Gabrielov1996ComplementsOS}, which contains all the sets of the form $\{(u,t)\in[-1,1]^n \times \Rm \mid  f(u) = t\}$ where $f\,:\,[-1,1]^n\rightarrow \Rm$ is an analytic function that can be analytically extended on a neighbourhood of $[-1,1]^n$. The second example is the \textit{$\log$-$\exp$ structure} $(\Rm_{\rm an}, \text{exp})$ \cite{dries_1998,Wilkie1996ModelCR}, which includes $\Rm_{\rm an}$ and the graph of the exponential function. Even though this second structure is a common setting for many optimisation problems, it does not meet the requirements for ours: as shown in \cite{logexp}, $\Gamma^{>0}$ (\textit{\emph{i.e.}\ }, the restriction of the Gamma function to $(0,+\infty)$) is not definable on $(\mathbb{R}_{\rm an},\exp)$. We thus consider the larger structure $(\mathbb{R_{\mathcal{G}}}, \exp)$, where $\Gamma^{>0}$ has been proved to be definable \cite{Dries2000TheFO}. $\mathbb{R_{\mathcal{G}}}$ is an o-minimal structure that extends $\mathbb{R}_{\rm an}$ and is generated by the class $\mathcal{G}$ of \textit{Gevrey} functions from \cite{Tougeron1994SurLE}.\\

{We end this section with the following result, which will be useful subsequently.\\

\begin{proposition}\label{prop:weakconvlogGam}
The function $t\mapsto\ln\Gamma(1+\frac{1}{t})$ defined on $(0,+\infty)$ is $\mu$-weakly convex
with $\mu > \mu_0 \approx 0.1136$.\\
\end{proposition}
\begin{proof}
Let us show that
there exists $\mu>0$ such that function $t\mapsto\ln\Gamma(1+\frac{1}{t}) + \mu t^2/2$ is convex
on $(0,+\infty)$.
The second-order derivative of this function on the positive real axis  reads
\begin{multline}\label{eq:loggamma:der2}
     \frac{d^2}{dt^2}\left(\ln\Gamma\left(1+\frac{1}{t}\right) + \frac{\mu}{2} t^2\right) =
  \\   \frac{1}{t^3}\left( 2\text{Dig}\left(1+\frac{1}{t}\right) +\frac{1}{t}\text{Dig}^{\prime}\left(1+\frac{1}{t}\right) +  \mu t^3\right),
\end{multline}
where the Digamma function $\operatorname{Dig}()$ is the logarithmic derivative of the Gamma function. In order to show the convexity of the considered function, 
we need to ensure that \eqref{eq:loggamma:der2} is positive for every $t\in(0,+\infty)$. By virtue of {Bohr–M\"ollerup}'s theorem {\cite[Theorem 2.1]{Artin1964TheGF}}, among all functions extending the factorial functions to the positive real numbers, only the Gamma function is log-convex. More precisely, its natural logarithm is (strictly) convex on the positive real axis. This implies that $t\mapsto\text{Dig}^{\prime}(t)$ is positive. It results that the only sign-changing term in \eqref{eq:loggamma:der2} is function $t \mapsto2\,\text{Dig}\left(1+\frac{1}{t}\right)$ as $t\mapsto \text{Dig}(t)$ vanishes in a point $t_0>1$ ($t_0\approx 1.46163$) which corresponds to the minimum point of the Gamma function -- and therefore also of its natural logarithm \cite{wrench}. As a consequence, the Digamma function is strictly positive for $t\in(t_0,+\infty)$, implying that $t\mapsto\text{Dig}\left(1+\frac{1}{t}\right)$ is 
strictly positive for all $t \in (0,\frac{1}{t_0-1} )$.  
Furthermore, $t\mapsto \text{Dig}\left(1+\frac{1}{t}\right)$ is strictly decreasing and bounded from below, as shown by the negativity of its first derivative
\begin{equation*}
\frac{d}{dt}\text{Dig}\left(1+\frac{1}{t}\right) = -\frac{1}{t^2}\text{Dig}^{\prime}\left(1+\frac{1}{t}\right)
\end{equation*}
and by the following limit
\begin{equation*}
    \lim_{t\rightarrow + \infty}    \text{Dig}\left(1+\frac{1}{t}\right) = \text{Dig}(1) = - \mathcal{E}
\end{equation*}
where the last equality holds by virtue of the Gauss Digamma theorem, and $\mathcal{E}$ is Euler-Mascheroni's constant $\mathcal{E}\approx 0.57721$ \cite{andrews_askey_roy_1999}. In conclusion, for $t \in [\frac{1}{t_0-1},+\infty)$, we need to ensure that the positive terms in \eqref{eq:loggamma:der2} manage to balance the negative contribution of function $t\mapsto 2 \text{Dig}\left(1 + \frac{1}{t}\right) > -2\mathcal{E}$. This leads to a condition on parameter $\mu>0$, since we can impose that 
$$0<\mu t^3 - 2\mathcal{E},$$
where the right-hand side expression has a lower bound
$\mu/(t_0-1)^3-2\mathcal{E}$
that is positive when
$$\mu  > 2{\mathcal{E}}(t_0-1)^{3} = \mu_0 \approx
0.1136.$$ 
This shows that function $t\mapsto\ln\Gamma(1+\frac{1}{t})$ is $\mu$-weakly convex. 
\end{proof}
}

\subsection{Proximal computations}

{    Let us now discuss the practical implementation of the proximal computations involved in \autoref{alg:ours}. Specifically, as we will show, none of these operators have closed-form expressions, so we need to resort to the inexact version. To ease the description, we summarise in \autoref{alg:ours:3} the application of \autoref{alg:ours} to the resolution of \eqref{eq:MAPNEGLOG}. 
{   As pointed out in \cite{Repetti2021VariableMetric} and in \cite{Bonettini2019}, the role of the \emph{relative error conditions} \eqref{eq:def:suff_dec} and \eqref{eq:def:inex_opt} are more of theoretical interest than of practical use. In the following, we will illustrate optimisation procedures ensuring that condition \eqref{eq:def:suff_dec} is satisfied for every block of variables at every iteration.}

\begin{centering}
\begin{algorithm}[htb]
 \textbf{Initialize} $x^0$, $p^0$ and $\beta^0$\;\\
  \textbf{Set} $\gamma_0 \in(0,1)$, $\gamma_1\in(0,1/\mu_0)$, $\gamma_2>0$\;\\
 \textbf{For} $\,\ell = 0,1,\ldots$\\ 
 \textbf{Set} ${A_{\ell}}\in \mathcal{S}_{n}$ \;\\
 \textbf{Find}\;
 {
 \begin{align}\label{eq:x}
  x^{\ell +1} &\approx \prox^A_{\gamma_0 q{(\cdot,p^{\ell},\beta^{\ell})}}(x^{\ell}-\gamma_0A^{-1}\nabla f(x^{\ell})) \\ &\qquad \text{\quad(with\;\autoref{alg:MM})}\notag\\
  \label{eq:p}
    p^{{\ell}+1} &\approx\prox_{\gamma_1\theta{(x^{{\ell}+1},\cdot,\beta^{\ell})}}(p^{{\ell}}) \\&\qquad \text{\quad (with\;\autoref{alg:CV})}\notag\\
    \label{eq:beta}
    \beta^{{\ell}+1} &\approx\prox_{\gamma_2\theta{(x^{{\ell}+1},p^{\ell+1},\cdot)}}(\beta^{\ell})\\&\qquad\text{\quad(with\;\autoref{alg:PD:2})}
    \notag\end{align}
 }
 \caption{P-SASL-PAM to solve \eqref{eq:MAPNEGLOG}}
 \label{alg:ours:3}
\end{algorithm}
\end{centering}}

\paragraph{Proximal computation with respect to $x$.} 
Subproblem \eqref{eq:x} in \autoref{alg:ours:3} requires the computation of the proximity operator of the following separable function $$q(\cdot, p^\ell,\beta^\ell)\,:\,x\,\mapsto \,\sum_{i=1}^{n} \left(C_\delta(x_i)\right)^{p_i^\ell}e^{-\beta_i^\ell p_i^\ell},$$
within a weighted Euclidean metric induced by some matrix $A\in\mathcal{S}_n$. 
 We notice that $x_i \mapsto \left(C_\delta(x_i)\right)^{p_i^\ell}$ is non-convex whenever $p^\ell_i \in (0,1)$, for some $i \in \{1,\dots,n\}$. In order to overcome this issue, we apply a majorisation principle \cite{MAJ}. 
{Let us introduce function $\sigma$ defined,
for every $u \in [\delta_1,+\infty)$, as  $\sigma(u) = (u-\delta_2)^{p}$ with $p\in (0,1]$,
and vector $\delta = (\delta_1,\delta_2)\in (0,+\infty)^2$ such that $\delta_2<\delta_1$.
Since this function is concave, it can be majorised by 
its first-order expansion around any
point $w >\delta_2$:
\begin{multline}
(\forall u > \delta_2) \quad
    (u-\delta_2)^{p}  \leq (w-\delta_2)^{p} \\ + p (w-\delta_2)^{p-1}(u-w)\\
 = (1-{p})(w-\delta_2)^{p} +{p}(w-\delta_2)^{p-1}(u-\delta_2).
\end{multline}
Setting, for every $(t,t') \in {\mathbb{R}^2}$, $ u = H_{\delta_1}(t) \geq \delta_1$, $w = H_{\delta_1}(t') \geq \delta_1$ allows us to deduce the following majorisation:}
\begin{align}
  ( C_\delta(t) )^p\leq 
    & (1-p)(C_\delta(t'))^p
    + p(C_\delta(t'))^{p-1}C_\delta(t) .\label{eq:majCd}
\end{align}
Let us now define $\mathcal{I}^\ell = \{i \in \{1,\dots,n\}\;|\;p^\ell_i\geq 1\}$
and  $\mathcal{J}^\ell =  \{1,\dots,n\}\setminus \mathcal{I}^\ell $.
Given $v=(v_i)_{1\le i \le n}\in\mathbb{R}^n $, we deduce from \eqref{eq:majCd} that
\begin{align}
    &(\forall x = (x_i)_{1\le i \le n} \in \mathbb{R}^n)\notag \\&\quad q(x,p^\ell,\beta^\ell) = \sum_{i\in\mathcal{I}^\ell} \left(C_{\delta}(x_i)\right)^{p^\ell_i}e^{-\beta^\ell_i p^\ell_i} \\&\qquad \qquad + \sum_{i\in\mathcal{J}^\ell} \left(C_{\delta}(x_i)\right)^{p^\ell_i}e^{-\beta^\ell_i p^\ell_i}\nonumber\\
    & \le  
\overline{q}(x,v,p^{\ell},\beta^{\ell}),
\end{align}
where the resulting majorant function is separable,
\textit{\emph{i.e.}\ }
\begin{equation}
    \overline{q}(x,v,p^{\ell},\beta^{\ell}),
= \sum_{i=1}^n 
   \overline{q}_i(x_i,v_i,p^{\ell}_i,\beta^{\ell}_i),
    \label{eq:q_J}
 \end{equation}
 with, for every $i \in \{1,\ldots,n\}$
 and $x_i \in \mathbb{R}$,
 \begin{align}
&\overline{q}_i(x_i,v_i,p^{\ell}_i,\beta^{\ell}_i)\\
     &=\begin{cases}
     \notag
 e^{-\beta^\ell_i p^\ell_i}\left(C_{\delta}(u_i)\right)^{p^\ell_i}, 
 &\mbox{if $p_i^\ell \ge 1$}\\
    e^{-\beta^\ell_i p^\ell_i}\Big((C_\delta(v_i))^{p^\ell_i}(1- p^\ell_i)   
    \\
    \qquad 
   + p^\ell_i(C_\delta(v_i))^{p^\ell_i-1}C_{\delta}(x_{i})\Big)
    & \mbox{otherwise.}
     \end{cases}
 \end{align}
 In a nutshell, each term of index $i \in \{1,\dots,n\}$ in \eqref{eq:q_J} coincides either with the $i$-th term of $q(\cdot,p^{\ell},\beta^{\ell})$ when $i\in \mathcal{I}^\ell$, or it is a convex majorant of this $i$-th term with respect to $v_i$
 when $i\in \mathcal{J}^\ell$.
We thus propose to adopt an MM procedure by building a sequence of convex surrogate problems for the non-convex minimisation problem involved in the computation of  $\prox_{\gamma_0 q(\cdot, p^\ell,\beta^\ell)}^A$. {At the $\kappa$-th iteration of this procedure,
following the MM principle, the next iterate $x^{\kappa+1}$ is determined by setting
$v = x^\kappa$.} We summarise the strategy in \autoref{alg:MM}.\\


{
\begin{algorithm}[htb]
 \textbf{Initialize} $x^0 \in 
 \Rmn$\;\\
\textbf{For} $\kappa = 0,1,\dots$ \textbf{until} convergence
{ \begin{align}
x^{\kappa+1} &= \prox_{\gamma_0 \overline{q}(\cdot, x^\kappa, p^{\ell},\beta^{\ell})}^{A}(x^+)\\
&\text{\quad(with\;\autoref{alg:DFB}) \notag}
  \end{align}
 }
 \caption{MM algorithm to approximate $\prox_{\gamma_0 q(\cdot, p^\ell,\beta^\ell)}^A(x^+)$ with $x^+\in \Rmn$}
 \label{alg:MM}
\end{algorithm}
}
Since function $\overline{q}(\cdot, v, p^{\ell},\beta^{\ell})$ is convex, proper, and lower semicontinuous, its proximity operator in the weighted Euclidean metric induced by matrix $A$ is guaranteed to be uniquely defined. It can be computed efficiently using the Dual Forward-Backward (DFB) method \cite{COMBETTES2011680}, outlined in \autoref{alg:DFB}.

\begin{algorithm}[htb]
 \textbf{Initialize} dual variable $w^{0}\in \mathbb{R}^n$\;\\
 \textbf{Set}  $\eta \in (0,2|||A|||^{-1})$
 \;\\
 \textbf{For} $\kappa' = 0,1,\dots$ \textbf{until} convergence{
 \begin{align}
 u^{\kappa'} &= x^+-Aw^{\kappa'},\\
  w^{\kappa'+1} &= w^{\kappa'} + \eta u^{\kappa'} \\
  &\;\;\notag- \eta \prox_{\eta^{-1}\gamma_0\overline{q}(\cdot,v,p^{\ell},\beta^{\ell})}(\eta^{-1}w^{\kappa'} + u^{\kappa'}).\label{eq:prox:qj}
  \end{align}
  \textbf{Return} $u^{\kappa^\prime} \in \mathbb{R}^n $
 }
 \caption{DFB algorithm to compute $\prox_{\gamma_0\overline{q}(\cdot,v; p^\ell,\beta^\ell)}^A(x^+)$ with $x^+\in \Rmn$}
 \label{alg:DFB}
\end{algorithm}
{

The update in \eqref{eq:prox:qj} can be performed componentwise since function
$\overline{q}(\cdot,v,p^{\ell},\beta^{\ell})$
is separable.
Thanks to the separability property,  
computing $\prox_{\eta^{-1}\gamma_0\overline{q}(\cdot,v,p^{\ell},\beta^{\ell})}$
boils down to solving $n$ one-dimensional optimization problems, that is
\begin{align}(\forall u^+ &= (u^+_i)_{1 \leq i \leq n} \in \Rmn)\nonumber\\
    &\prox_{\eta^{-1}\gamma_0\overline{q}(\cdot,v,p^{\ell},\beta^{\ell})}(u^+)\nonumber\\
    &= \left( \prox_{\eta^{-1}\gamma_0\overline{q}_i(\cdot,v_i,p^{\ell}_i,\beta^{\ell}_i)}(u^+_i)\right)_{1\leq i\leq n}.
\end{align}

More precisely, 
\begin{itemize}
    \item for every $i \in \{1,\ldots,n\}$, such that $p^\ell_i \le 1$,
  \begin{align}
&\prox_{\eta^{-1}\gamma_0\overline{q}_i(\cdot,v_i,p^{\ell}_i,\beta^{\ell}_i)}(u^+_i) \nonumber\\
&=\prox_{\eta^{-1}\gamma_0e^{-\beta^\ell_i p^\ell_i} p^\ell_i(C_\delta(v_i))^{p^\ell_i-1} C_{\delta_1}}(u^+_i) \nonumber\\
  &=\prox_{\eta^{-1}\gamma_0e^{-\beta^\ell_i p^\ell_i} p^\ell_i(C_\delta(v_i))^{p^\ell_i-1} H_{\delta_1}}(u^+_i).
    \end{align}
    The proximity operator of the so-scaled version of function $H_{\delta_1}$ can be determined by solving a quartic polynomial equation.\footnote{
    http://proximity-operator.net/scalarfunctions.html}
    \item For every $i\in \{1,\ldots,n\}$ such that $p^\ell_i > 1$,
   \begin{multline}
\prox_{\eta^{-1}\gamma_0\overline{q}_i(\cdot,v_i,p^{\ell}_i,\beta^{\ell}_i)}(u^+_i) \\= 
  \prox_{\eta^{-1}\gamma_0e^{-\beta^\ell_i p^\ell_i}
  \left(C_{\delta}\right)^{p^\ell_i}}(u^+_i).
    \end{multline}
    The latter quantity can be evaluated through a bisection search to find the root of the derivative of the involved proximally regularised function.
\end{itemize}}

\begin{remark}
{Due to the non-convexity of $q(\cdot, p^\ell,\beta^\ell)$, there is no guarantee that 
the point estimated by \autoref{alg:DFB}
coincides with the exact proximity point.
However, 
we did not notice any numerical issues in our 
implementation.}
\end{remark}

\paragraph{Proximal computation with respect to $p$.}

Subproblem \eqref{eq:p} requires to compute the proximity operator of $\gamma_1\big(q(x^{\ell+1},\cdot,\beta^\ell) + g\big)$,
which is equivalent to solving the following minimization problem
\begin{equation}
    \minimize{p\in 
    {[a,b]^n}}{} \psi^{\ell}(p) + \lambda\ell_{1,2}(Dp),
    \label{eq:qp_split}
    \end{equation}
where, for every $p \in \Rmn$, $\psi^{\ell}(p) = \sum_{i=1}^n \psi_i^{\ell}(p_i)$ with
\begin{multline}
(\forall i \in \{1,\ldots,n\})(\forall p_i \in \mathbb{R})\\ 
\psi_i^{\ell}(p_i)
= 
\begin{cases}
\left(C_{\delta}(x_i^{\ell+1})\right)^{p_i}e^{-\beta_i^{\ell}p_i} + \ln\Gamma(1+\frac{1}{p_i}) \\ \quad+ \frac{1}{2\gamma_1}(p_i-p_i^\ell)^2
\quad  \mbox{if $p_i > 0$}\\
+\infty \qquad\qquad\qquad \;\; \mbox{otherwise.}
\end{cases}
\end{multline}
{Moreover, $D = [D_{\rm h},D_{\rm v}]$ where $(D_{\rm h},D_{\rm v})\in (\mathbb{R}^{n\times n})^2$ are the discrete horizontal and vertical 2D gradient operators, and the $\ell_{1,2}$-norm is defined as
$$(\forall p \in \Rmn)\quad  \ell_{1,2}(D p) = \sum_{i=1}^n \| ([D_{\rm h} p]_i,[D_{\rm v} p]_{i})\|_2.
$$

}

{Problem~\eqref{eq:qp_split} is equivalent to minimizing the sum of the indicator function of a hypercube, a separable component and a non-separable term involving the linear operator $D$. According to Proposition \ref{prop:weakconvlogGam},  we can ensure the convexity of each term $(\psi_i^\ell)_{1\le i \le n}$ by 
setting $\gamma_1 < \frac{1}{\mu_0} \approx 8.805$.   In order to solve \eqref{eq:qp_split}, it is then possible to implement a Primal-Dual (PD) algorithm \cite{CONDAT,Duality,Vu2013ASA}  as outlined in \autoref{alg:CV}.\\

\begin{centering}
\begin{algorithm}[htb]
\textbf{Initialise} the dual variables $v^0_1\in \mathbb{R}^{n\times 2} $,$v^0_2\in \mathbb{R}^{n} $.\\
\textbf{Set} $\tau>0$ and $\sigma>0$ such that $\tau\sigma(|||D |||^2+1)< 1$.\\
\textbf{for} $\kappa = 0,1,\dots$ \textbf{until} convergence
\begin{align}
    u^{\kappa} &=  p^{\kappa} - \tau(D^*v^{\kappa}_1 + v^{\kappa}_2),\\
    {p}^{\kappa+1} &= \text{proj}_{[a,b]^n}(u^{\kappa}),\\
    w^{\kappa}_1 &=v^{\kappa}_1 + \sigma D(2{p}^{\kappa+1}- {p}^{\kappa}),\\
    v^{\kappa+1}_1 &= w^{\kappa+1}_1 - \sigma\text{prox}_{\frac{\lambda\ell_{1,2}}{\sigma}}(\frac{w^{\kappa}_1}{\sigma}).\\
    w^{\kappa}_2 &=v^{\kappa}_2 + \sigma (2{p}^{\kappa+1}- {p}^{\kappa}),\\
    v^{\kappa+1}_2 &= w^{\kappa+1}_2 - \sigma\text{prox}_{\frac{\psi^\ell}{\sigma}}(\frac{w^{\kappa}_2}{\sigma}).\label{eq:CV:v2}
\end{align}
\textbf{Return} $ p^{\kappa+1} \in [a,b]^n$
\caption{Primal Dual Algorithm for solving \eqref{eq:qp_split}}
 \label{alg:CV}
\end{algorithm}
\end{centering}

The proximity operator of the involved $\ell_{1,2}$ norm has a closed-form expression{. For every $w_1 = ([w_1]_{i,1},[w_1]_{i,2})_{1 \leq i \leq n} \in \mathbb{R}^{n\times 2}$ and $\lambda>0$, we have 
\begin{align*}
    &\prox_{\lambda \ell_{1,2}}(w_1) \\
    &\quad= \left( \prox_{\lambda \|\cdot\|_2}\Big(([w_1]_{i,1},[w_1]_{i,2})\Big) \right)_{1 \leq i \leq n}\\
    &=\Bigg(([w_1]_{i,1},[w_1]_{i,2}) \\& \qquad - \frac{\lambda ([w_1]_{i,1},[w_1]_{i,2})}{\max\{\lambda, \|([w_1]_{i,1},[w_1]_{i,2})\|_2\}} \Bigg)_{1 \leq i \leq n}.
\end{align*}} 

The proximal point at ${{w_2^\kappa}/{ \sigma} = \left({[w_2^\ell]_i}/{ \sigma} \right)_{1\leq i\leq n}\in \Rmn}$ of the separable term $\psi^{\ell}$ with respect to a step size $1/\sigma$ can be found by minimizing, for every $i\in \{1,\ldots,n\}$, the following smooth function 
$$(\forall t \in (0,+\infty)) \quad \mathsf{g}_{1,i}(t) =
\psi_i^\ell (t)
+ \frac{\sigma}{2}\Big(t - \frac{[w_2^\kappa]_i}{ \sigma}\Big)^2.$$
 {The update in \eqref{eq:CV:v2} then reads $$v_2^{\kappa+1} = \left([w_2^{\kappa+1}]_i - \sigma [ w_2^{\kappa}]_i^* \right)_{1 \leq i \leq n}$$
 where, for every $ i \in \{1,\dots,n\}$,  $[ w_2^{\kappa}]_i^*$ corresponds to the unique zero of the derivative of $\mathsf{ g}_{1,i}$. This zero is found by applying Newton's method initialised with 
 $$\bar{w}_i = \left(\max\Big\{10^{-3},\frac{[w_2^\kappa]_i}{\sigma}\Big\}\right)_{1\leq i\leq n}.$$
 }

  }

\paragraph{Proximal computation with respect to $\beta$.}

Subproblem \eqref{eq:beta} requires the solution of the following minimisation problem:
 \begin{equation}
    \minimize{\beta\in\mathbb{R}^{n}} {\varphi^{\ell}(\beta) + \zeta \ell_{1,2}(D\beta)}
     \label{eq:beta:compact}
\end{equation}    
where $D$ and $\ell_{1,2}$ have been defined previously and
$$(\forall \beta = (\beta_i)_{1\le i \le n} \in \Rmn) \quad \varphi^{\ell}(\beta) = \sum_{i=1}^n \varphi_{i}^{\ell}(\beta_i)$$
with, for every $i \in \{1,\ldots,n\}$,
\begin{multline}
\varphi_{i}^{\ell} (\beta_i) = \left(C_{\delta}(x_i^{\ell+1})\right)^{p_i^{\ell+1}}e^{-\beta_i p_i^{\ell+1}} + \beta_i \\+  { \frac{(\beta_i-\mu_{\beta})^2}{2\sigma_\beta^2}}+ \frac{1}{2\gamma_2}(\beta_i-\beta_i^\ell)^2 \end{multline}
The above problem shares a structure similar to the one studied in the previous case since the objective function
is the sum of the smooth convex term $\varphi^{\ell}$
and the non-smooth convex one $\zeta \operatorname{TV} = \zeta\ell_{1,2}(D\cdot)$, 
and it can be solved by the primal-dual procedure
outlined in \autoref{alg:PD:2}. 

\begin{centering}
\begin{algorithm}[htb]
\textbf{Set} $\tau>0$ and $\sigma>0$ such that $ \tau\sigma|||D|||^2\leq  1$.\\
\textbf{Initialise} the dual variable $v^0\in \mathbb{R}^{n\times 2}$.\\
\textbf{for} $\kappa = 0,1,\dots$ \textbf{until} convergence
\begin{align}
    u^{\kappa} &=  \beta^{\kappa} - \tau(D^*v^{\kappa}),\\
    {\beta}^{\kappa+1} &= \text{prox}_{\tau\varphi^{\ell}}(u^{\kappa}),\label{eq:PD:beta}\\
    w^{\kappa} &=v^{\kappa} + \sigma D(2{\beta}^{\kappa+1}- {\beta}^{\kappa}),\\
    v^{\kappa+1} &= w^{\kappa+1} - \sigma\text{prox}_{\frac{\zeta \ell_{1,2}}{\sigma}}(\frac{w^{\kappa}}{\sigma}).
\end{align}
\textbf{Return} $\beta^{\kappa+1} \in \mathbb{R}^n$
\caption{Primal Dual Algorithm for minimizing \eqref{eq:beta:compact}}
 \label{alg:PD:2}
\end{algorithm}
\end{centering}

At each iteration $\kappa$ of \autoref{alg:PD:2}, the proximity operator of $\varphi^\ell$ 
is expressed as
\begin{multline}
(\forall \beta = (\beta_i)_{1 \leq i \leq n} \in \Rmn)\\
    \prox_{\tau\varphi^\ell}(\beta) = \left(\prox_{\tau\varphi_i^\ell}(\beta_i) \right)_{1\leq i \leq n}.
\end{multline}
For every $i \in \{1,\ldots,n\}$,
$\prox_{\tau\varphi_i^\ell}(\beta_i)$ is the minimizer of function
\begin{align}
(\forall \beta_i \in \mathbb{R}) \quad
     \mathsf {g}_{2,i}(\beta_i) = \varphi_i^{\ell}(\beta_i)+ \frac{1}{2\tau}(\beta_i-u_i^\kappa)^2.
\end{align}
The nonlinear equation defining the unique zero of the derivative of $\mathsf {g}_{2,i}$ admits a closed-form solution that involves the Lambert $W$-function \cite{Corless96onthe}. 
 Indeed, let us introduce the following notation: 
 {\allowdisplaybreaks\begin{align}
     a_{1,i} &=  p_i^{\ell+1}\left(C_{\delta}(x_i^{\ell+1})\right)^{p_i^{\ell+1}},\\
      a_{2} & = \left(\frac{1}{\sigma^2_{\beta}} + \frac{1}{\gamma_2} + \frac{1}{\tau}\right)^{-1},\\
      a_{3,i} &= 1 {  - \frac{\mu_{\beta}}{\sigma^2_{\beta}} }- \frac{\beta_i^\ell}{\gamma_2} - \frac{u_i^\kappa}{\tau}.
 \end{align}}
Then 
 \begin{align}
&\mathsf {g}_{2,i}^{\prime}(\beta_i) = 0 \notag\\
&\iff - a_{1,i} \exp(-p_i^{\ell+1} \beta_i) + \frac{\beta_i}{a_2} + a_{3,i} = 0\nonumber\\ 
&\notag \iff p_i^{\ell+1} (\beta_i + a_2 a_{3,i})\exp(p_i^{\ell+1} (\beta_i + a_2 a_{3,i}))\\ &\qquad \qquad = p_i^{\ell+1} a_{1,i} a_2 \exp(p_i^{\ell+1} a_2 a_{3,i})\nonumber\\
&\notag \iff \beta_i \\&= \frac{1}{p_i^{\ell+1}} W( p_i^{\ell+1} a_{1,i} a_2 \exp(p_i^{\ell+1} a_2 a_{3,i}) ) - a_2 a_{3,i},
\label{eq:step3:prox}
 \end{align}
 where the last equivalence comes from the fact that the Lambert $W$-function is single-valued and satisfies the following identity for a pair {  $(X,Y)\in \mathbb{R}\times\left(-\frac{1}{e},+\infty\right)$}:
 \begin{equation}
    X\exp(X) = Y \iff X = W(Y). 
 \end{equation}
 {  Notice that the expression in \eqref{eq:step3:prox} is well defined since the argument of the Lambert function is always positive.}
 
 {In conclusion, the update in \eqref{eq:PD:beta}
 reads as $\beta^{\kappa+1} = \left(\beta_i^{\kappa+1}\right)_{1 \leq i \leq n}$ where each component of this vector is calculated according to \eqref{eq:step3:prox}. \\
 }

 { 
 \section{Numerical Experiments}\label{sec:simulation}

}{ 

In this section, we illustrate the performance of our approach on a problem of joint deblurring/segmentation of realistically simulated ultrasound images. We consider images with two regions (\emph{Simu1}) and three regions (\emph{Simu2}) extracted from \cite{PPULA}. {Both images have dimension $256\times256$ pixels. The shape parameters $p$ and the reparameterised scale parameters $\beta$ are set in each region following the choices for $p$ and $\alpha$ in \cite{PPULA}, itself based on the experimental setting in \cite{US_GGD}. This strategy allows us to have a reference configuration for $\beta$, which led us to choose a non-necessarily zero-mean Gaussian distribution as a prior for this parameter. In our experiments, we will treat $\mu_\beta$ as an unknown parameter, along with the regularisation parameters for the Total Variation priors. The pixel values in each region of the original image $x\in\Rm^n$ are obtained as a realisation of a random variable following a $\mathcal{GGD}$ with the corresponding shape and scale parameters $p$ and $\alpha$.} We define $K$ as the linear operator modelling the convolution with the point spread function of a 3.5 MHz linear probe obtained with the Field II ultrasound simulator \cite{Jensen2004Simulations}. 
 To reproduce the same setting as in \cite{PPULA}, we obtain the observed degraded images $y\in\Rm^n$ from the original images $x\in\Rm^n$ by applying the observation model \eqref{eq:model}, where we set the additive noise variance (which will be assumed to be known) to $\sigma^2 = 0.013$ for \emph{Simu1} and $\sigma^2 = 33$ for \emph{Simu2}. For the preconditioner, we consider a regularised version of the inverse of the Hessian of the data fidelity function in \eqref{eq:f}, given by
 $$A = \sigma^2(K^{\top}K + \mu \mathbb{I}_m)^{-1}$$ where $\mu = 0.1$, so that $A$ is well defined and constant throughout the iterations.
Following the procedure outlined in \cite{PPULA}, we initialise ${x}^0\in\Rm^n$ using a pre-deconvolved image obtained with a Wiener filter applied to the observed data $y$, $({p}^0_i)_{1\le i \le n}$ is drawn from an i.i.d.\ uniform distribution in the range $[0.5,1.5]$, while $({\beta}^0_i)_{1\le i \le n}$ is drawn from an i.i.d.\ Gaussian distribution with mean  $\mu_{\beta}$ and unit standard deviation. We set $\mu_\beta = 0$ for \textit{Simu 1} and $\mu_\beta = 4$ for \textit{Simu 2}, for arguments discussed in SM 2. We adopt the recovery strategy described in \autoref{sec:application} and describe hereafter the setting of the model/algorithm hyperparameters.\\

The model parameters that need to be tuned are the $\delta_1>0$ and $\delta_2>0$ values for the pseudo-Huber function, the mean $\mu_{\beta}\in \R$ and the standard deviation $\sigma_{\beta}>0$ for the reparameterised scale parameter, and finally the regularisation parameters $(\lambda,\zeta) \in (0,+\infty)^2$ for the TV terms. 
For parameter $\delta=(\delta_1,\delta_2)$, we applied the following choice, resulting from a rough empirical search,: $\delta_1 = 1$  while $\delta_2 = \delta_1\times 10^{-2}$. For what concerns the Gaussian parameters of the reparameterised scale variable $(\mu_{\beta}, \sigma_\beta)$, the mean $\mu_{\beta}$ is the most influential on the estimated solution, so we dedicated an in-depth analysis for its choice in combination with the TV regularisation parameters $(\lambda,\zeta)$.  
More precisely, we tested different values of $\mu_{\beta}$ in the range $[-10,10]$ in combination with a grid search for $(\lambda,\zeta) \in \{10^{-2},10^{-1}, 1, 10, 10^2, 10^3\}^2$ with respect to different quality metrics and identified an optimal choice for $\mu_{\beta}$. The standard deviation appeared less influential and is set to $\sigma_{\beta}=1$ in all our experiments. The details of the analysis are illustrated in the annexed SM 2.

  The algorithmic hyperparameters include the step sizes of the proximal steps, as well as the preconditioning matrix involved in the preconditioned proximal gradient step.  We set $(\gamma_0,\gamma_1,\gamma_2) = (0.99,1,1)$
in order to meet the convergence assumptions in \autoref{alg:ours:3}. In particular, the choice for $\gamma_0$ approximates the highest value allowed for the step size of the preconditioned inexact FB scheme in \eqref{eq:x}, while $\gamma_1$ satisfies the condition $\gamma_1 < 8.805$ for the convexity of the function in \eqref{eq:qp_split}.

In order to obtain the labelling of a segmented image from our estimated shape parameter (denoted by $\widehat p$) we use a quantisation procedure based on Matlab functions \texttt{multithresh} and \texttt{imquantize}. The former defines a desired number of quantisation levels using Otsu’s method, while the latter performs a truncation of the data values according to the provided quantisation levels. We remark here that the number of labels does not need to be defined throughout the proposed optimisation procedure, but only at the final segmentation step. This step can thus be considered as a post-processing that is performed on the estimated solution.

 In order to evaluate the quality of the solution, we consider the following metrics: for the estimated image, we make use of  the peak signal-to-noise ratio (PSNR) defined as follows, $x$ being the original signal and $\widehat{x}$ the estimated one: 
 \begin{equation*}
     \text{PSNR} = 10\log_{10}\big(n\,\max_{i\in\{1,\dots,n\}} \{x_i,\hat{x}_i\}^2/\|x_i-\hat{x}_i\|^2\big),
 \end{equation*} and of the structure similarity measure (SSIM) \cite{ssim}. For the segmentation task we compute the percentage OA of correctly predicted labels.
 
 The stopping criteria for \autoref{alg:ours:3} outer and inner loops are set by defining a threshold level on the relative change between two consecutive iterates of the involved variables, the relative change of the objective values of two consecutive iterates and a maximum number of iterations. The outer loop in \autoref{alg:ours:3} stops whenever $\ell=10000$ or when both $\|\zeta^{\ell + 1} - \zeta^\ell\|/\|\zeta^\ell\| < 10^{-4}$ and $|\theta(\zeta^{\ell + 1}) - \theta(\zeta^\ell)|/|\theta(\zeta^\ell)| < 10^{-4}$. The MM procedure to compute $x^{\ell+1}$ in \autoref{alg:MM} is stopped after 300 iterations or when $\|x^{\kappa + 1} - x^\kappa\|/\|x^\kappa\| < 10^{-3}$. The DFB procedure in \autoref{alg:DFB} to compute $u^{\kappa+1}$ is stopped after 300 iterations or when $\|u^{\kappa + 1} - u^\kappa\|/\|u^\kappa\| < 10^{-3}$. The PD procedure in \autoref{alg:CV}, and \autoref{alg:PD:2} computing $p^{\ell+1}$ (resp. $\beta^{\ell+1}$) terminates after 200 iteration or when $\|p^{\kappa + 1} - p^\kappa\|/\|p^\kappa\| < 10^{-3}$ (resp. $\|\beta^{\kappa + 1} - \beta^\kappa\|/\|\beta^\kappa\| < 10^{-3}$). \\

\autoref{fig:simu1_and_simu2} illustrates in the first and second line the B--mode image of the original $x$, of the degraded $y$, and of the reconstructed image $\hat{x}$ on both examples. The B--mode image is the most common representation of an ultrasound image, displaying the acoustic impedance of a 2-dimensional cross section of the considered tissue. The reconstructed results in \autoref{fig:simu1_and_simu2} (right) show clearly reduced blur and sharper region contours. We then report in the third and fourth lines of \autoref{fig:simu1_and_simu2}  the estimated shape parameter and the segmentation obtained via the aforementioned quantisation procedure, which confirms its good performance. 
 We notice that our estimated $\hat{p}_i$ values are consistent with the original ones and the fact that the results for \emph{Simu2} are slightly less accurate than the ones for \emph{Simu1} is in line with the results presented in \cite[Table III]{PPULA} for P-ULA, HMC and PP-ULA, suggesting that the configuration of the parameters for \emph{Simu2} is quite challenging.
 

\autoref{tab:simu1_and_simu2:metrics} proposes a quantitative comparison of our results against those of the methods considered in \cite{PPULA}: a combination of Wiener deconvolution and Otsu's segmentation \cite{otsu}, a combination of LASSO deconvolution and SLaT segmentation \cite{SLAT}, the adjusted Hamiltonian
Monte Carlo (HMC) method \cite{hmc}, the Proximal Unadjusted Langevin algorithm (P-ULA) \cite{pula} and its preconditioned version (PP-ULA) \cite{PPULA} for joint deconvolution and segmentation. 
From this table, we can conclude that the proposed variational method is able to compete with state-of-the-art Monte Carlo Markov Chain techniques in terms of both segmentation and deconvolution performance. For what concerns the computational time,  the average time (over 10 runs of the algorithm) required by P-SASL-PAM to meet the stopping criteria ${\|\zeta^{\ell + 1} - \zeta^\ell\|/\|\zeta^\ell\| < 10^{-4}}$ and ${|\theta(\zeta^{\ell + 1}) - \theta(\zeta^\ell)|/|\theta(\zeta^\ell)| < 10^{-4}}$ corresponds to $493.2$ seconds (approximately $8' 13{''}$) for \emph{Simu1}  and $536.4$ seconds (approximately $8'56{''}$) for \emph{Simu2}. Simulations were run on Matlab 2021b on an Intel Xeon Gold 6230 CPU 2.10GHz. 
 In \autoref{tab:simu1_and_simu2:metrics}, we report the computational times for PULA, HMC and PP-ULA from {\cite[TABLE II]{PPULA}, which were obtained on Matlab 2018b on an Intel Xeon CPU E5-1650 3.20GHz.

\begin{table*}[h]
    \centering
    \resizebox{1\textwidth}{!}{
    \begin{tabular}{ccccccccc}
   \toprule
    \multicolumn{1}{c}{} & \multicolumn{4}{c}{\emph{Simu1}}  & \multicolumn{4}{c}{\emph{Simu2}} \\
    \cmidrule(r){2-5}\cmidrule(r){6-9}
    METHOD   & PSNR & SSIM   & OA & TIME & PSNR & SSIM   & OA & TIME\\
   \cmidrule(r){1-1} \cmidrule(r){2-5}\cmidrule(r){6-9}
      Wiener-Otsu & 37.1 & 0.57 & 99.5 & -- &35.4 & 0.63 &  96.0 & --\\
      LASSO-SLaT & 39.2 & 0.60 & 99.6 & -- & 37.8 & 0.70 & 98.3 & --\\
      \cmidrule(r){1-1} \cmidrule(r){2-5}\cmidrule(r){6-9}
      P-ULA & 38.9 & 0.45 &98.7 & $2$ h $27$ min & 37.1 & 0.57 &94.9 & $3$ h $06$ min\\
      HMC & 40.0 & 0.62  & 99.7& $1$ h $08$ min &  36.4 & 0.64 & 98.5 & $4$ h $14$ min\\
      PP-ULA & 40.3 & 0.62  & 99.7& $12$ min & 38.6 & 0.71 & 98.7 & $39$ min\\
      \cmidrule(r){1-1} \cmidrule(r){2-5}\cmidrule(r){6-9}
      OURS & 40.2 & 0.61 & 99.9 & $8$ min& 38.1 & 0.70  & 97.7 &  $9$ min\\
     \bottomrule
    \end{tabular}}
    \caption{PSNR, SSIM, OA scores and Computational time for \emph{Simu1} and \emph{Simu2} from \cite{PPULA}. The symbol "--" means the result was not available in the reference paper.}\label{tab:simu1_and_simu2:metrics}
\end{table*}

 \begin{figure*}[h!]
     \centering
     \begin{tabular}{cccc}
     \toprule
      &  ORIGINAL & DEGRADED & RECONSTRUCTED\\
      \hline\\
     \rotatebox{90}{\emph{Simu1}}& 
     \makecell{\includegraphics[width=1.2in]{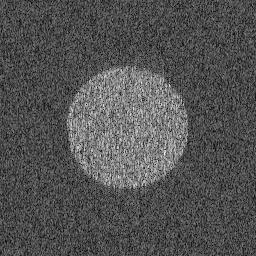}}&\makecell{\includegraphics[width=1.2in]{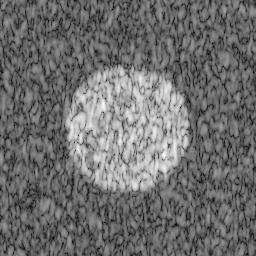}}&\makecell{\includegraphics[width=1.2in]{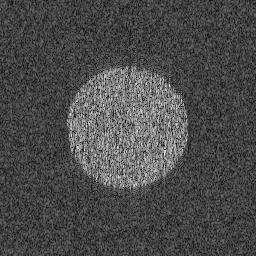}}\\
     \rotatebox{90}{\emph{Simu2}}&
 \makecell{\includegraphics[width=1.2in]{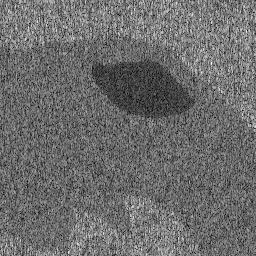}}&\makecell{\includegraphics[width=1.2in]{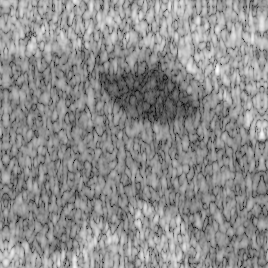}}&\makecell{\includegraphics[width=1.2in]{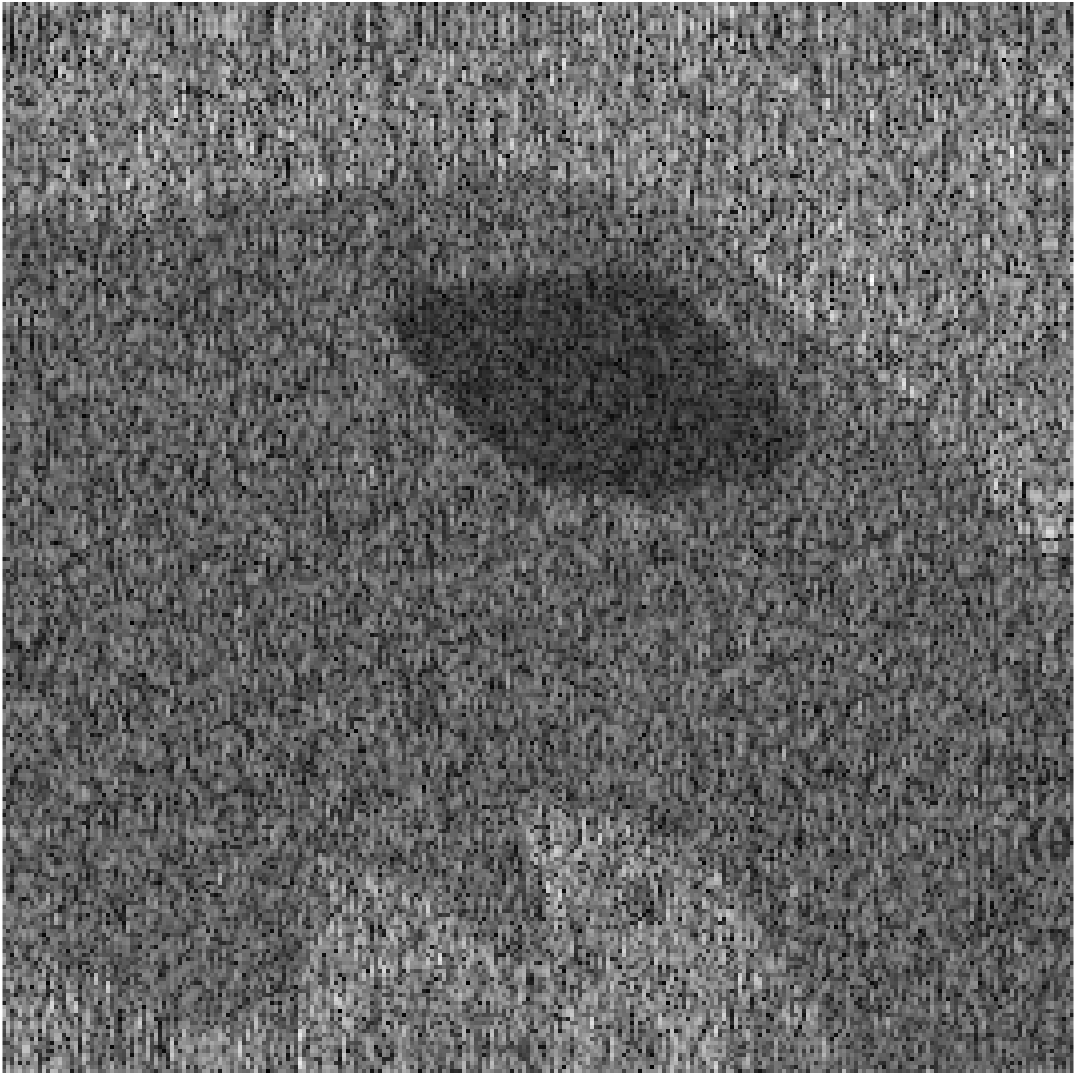}}\\
  
     \toprule
   & REFERENCE & ESTIMATED & QUANTISED \\ 
   \hline \\\makecell{\rotatebox{90}{\emph{Simu1}}} & 
     \makecell{\includegraphics[height=1.2in]{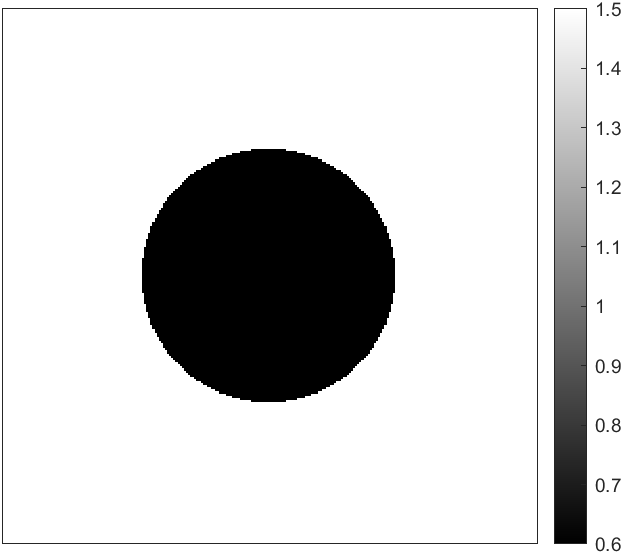}}&\makecell{\includegraphics[height=1.2in]{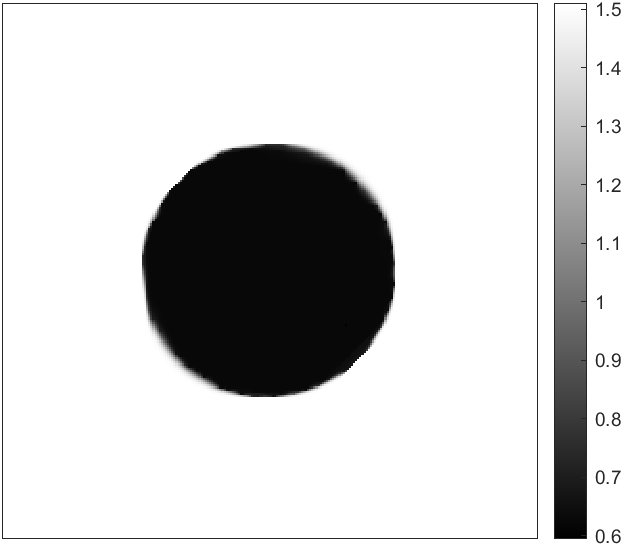}} & \makecell{\includegraphics[height=1.2in]{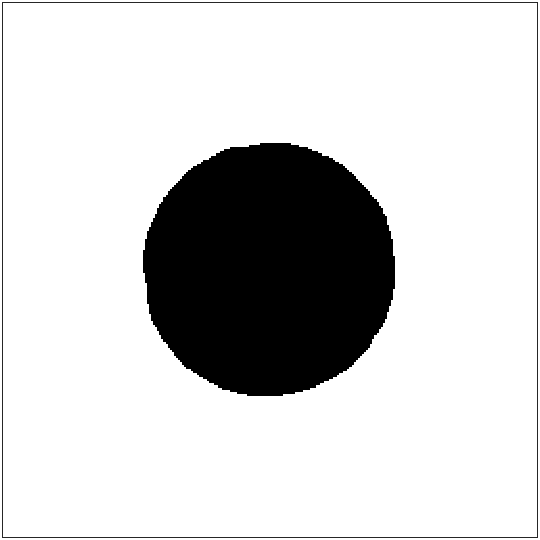}}
     \\
   \makecell{\rotatebox{90}{\emph{Simu2}}}  & 
     \makecell{\includegraphics[height=1.2in]{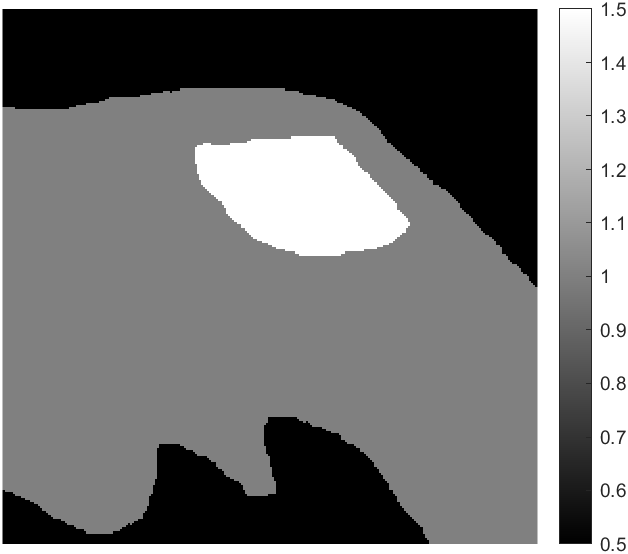}}&
     \makecell{\includegraphics[height=1.2in]{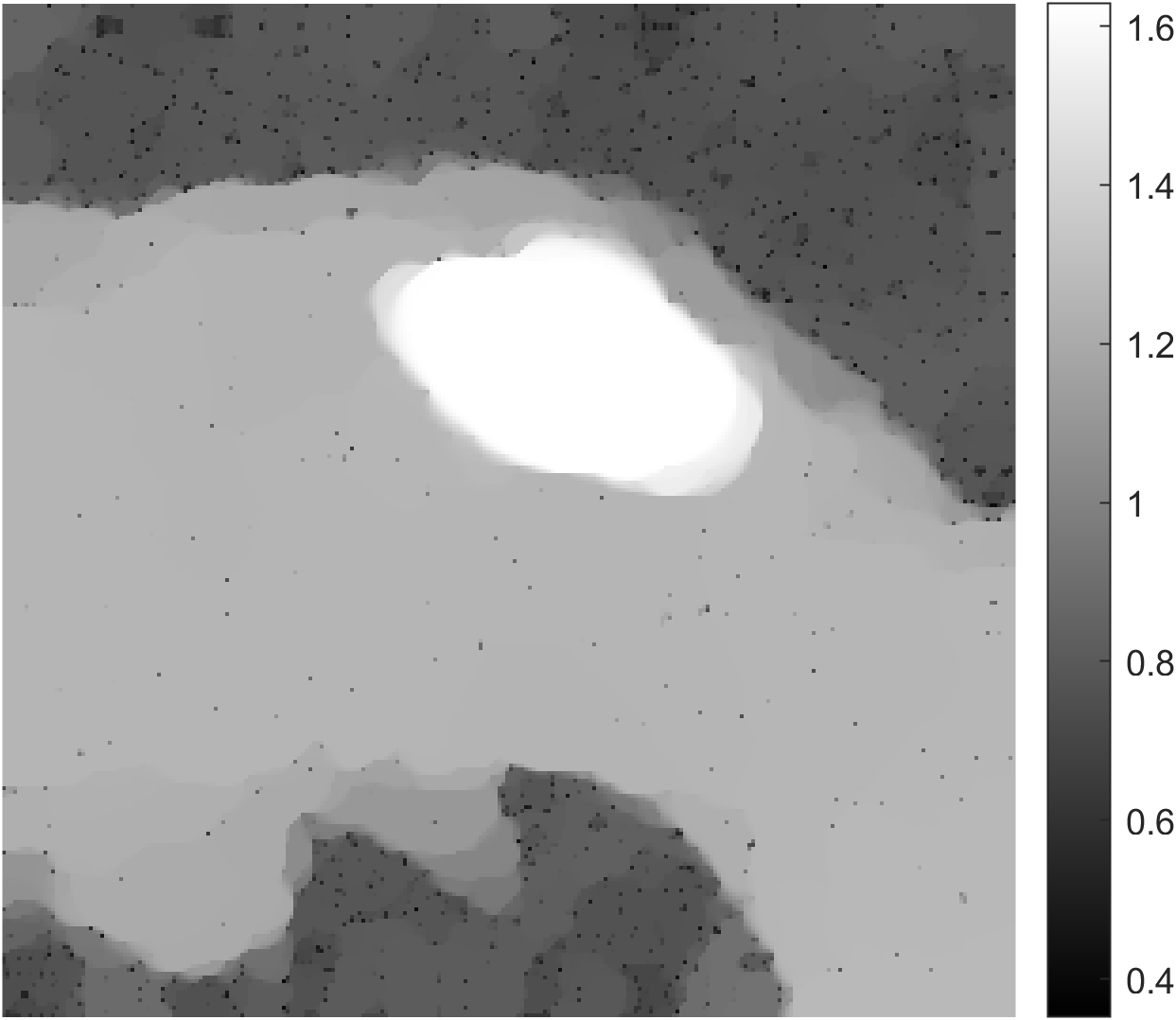} }
     & \makecell{\includegraphics[height=1.2in]{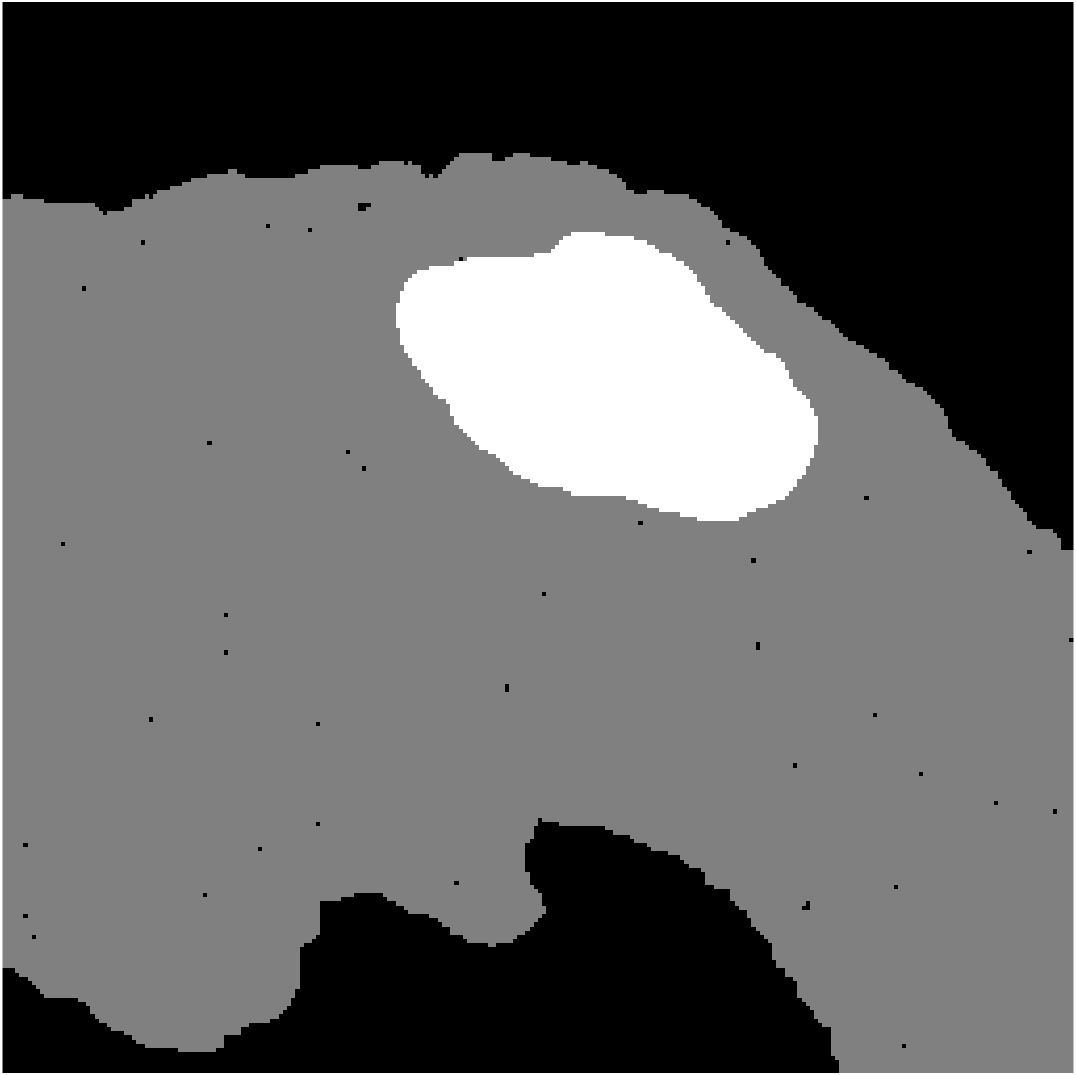} }
     \\
\bottomrule
     \end{tabular}
     \caption{First and Second lines: B--mode of Simu1 and Simu2. The B--mode image is the most common type of ultrasound image, displaying the acoustic impedance of a 2-dimensional cross section of the considered tissue. All images are presented in the same scale [0,1]. Third and Fourth lines: Segmentation of the shape parameter for \emph{Simu1} and \emph{Simu2}: reference $p$, estimated $\hat{p}$ and quantised $\bar{p}$.}
     \label{fig:simu1_and_simu2}
 \end{figure*}

\begin{figure*}
\resizebox{0.99\textwidth}{!}{
   \centering
     \begin{tabular}{cccc}
     \includegraphics[height = 1.025in]{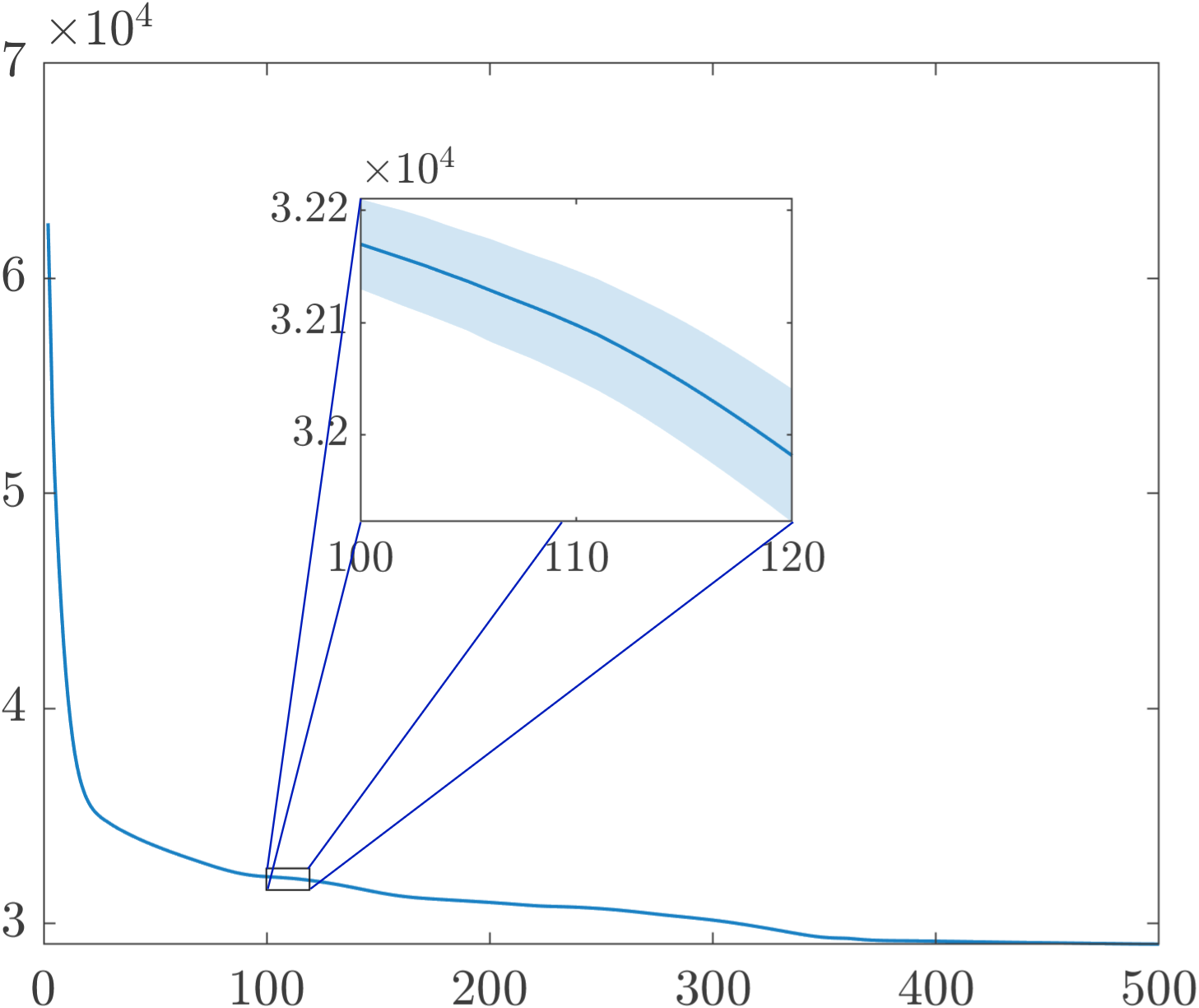}
     &\includegraphics[height = 1.03in]{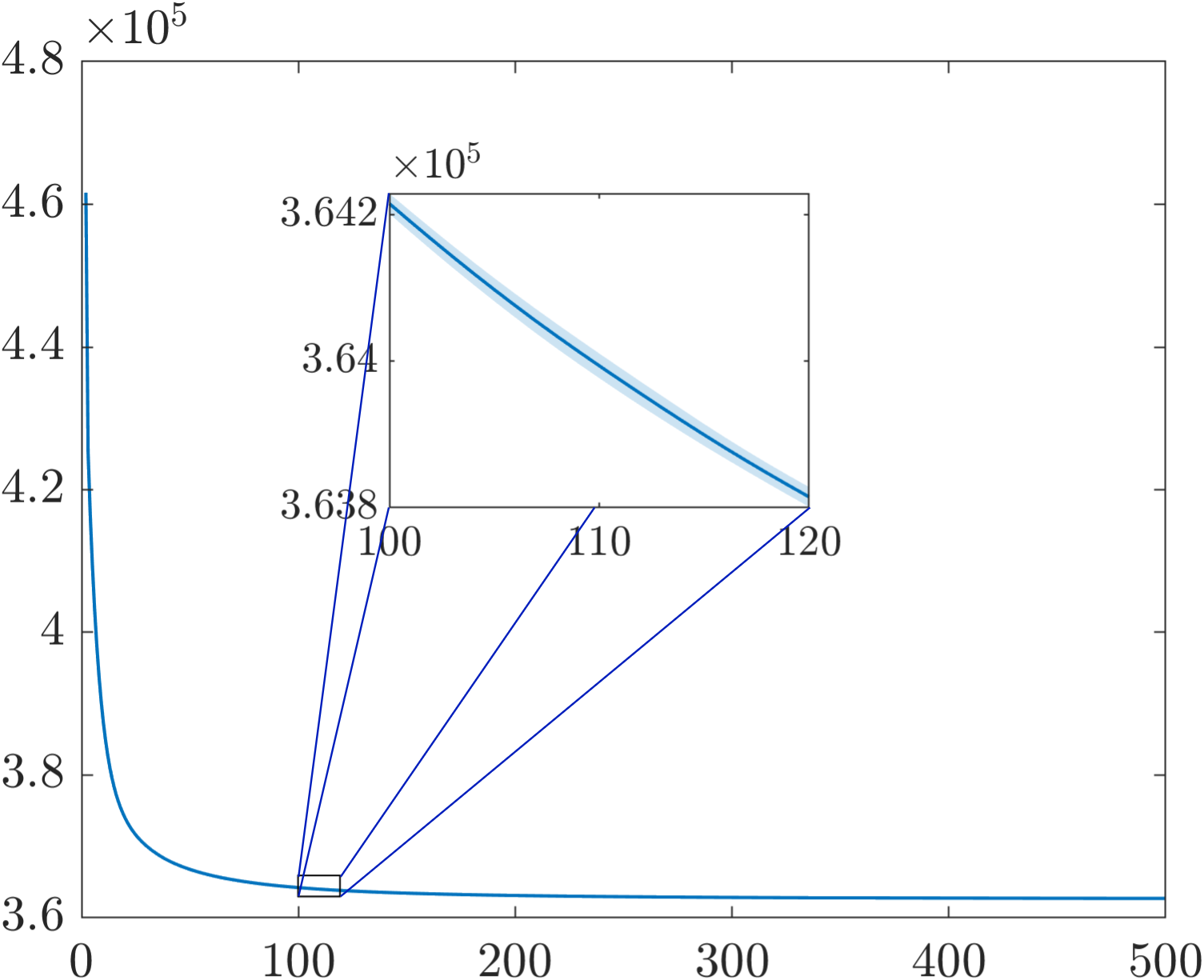} & \includegraphics[height = 1in]{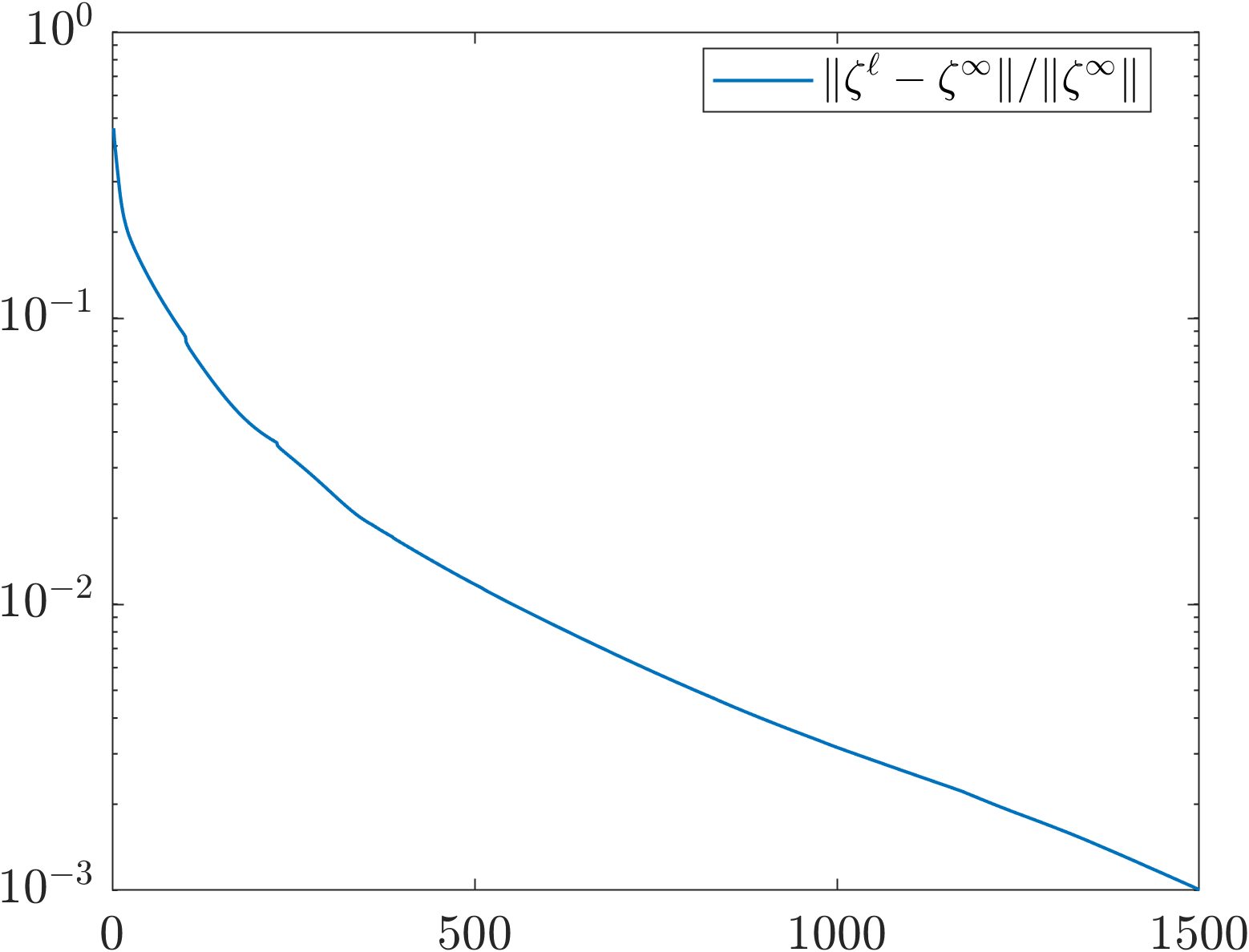}&
     \includegraphics[height = 1in]{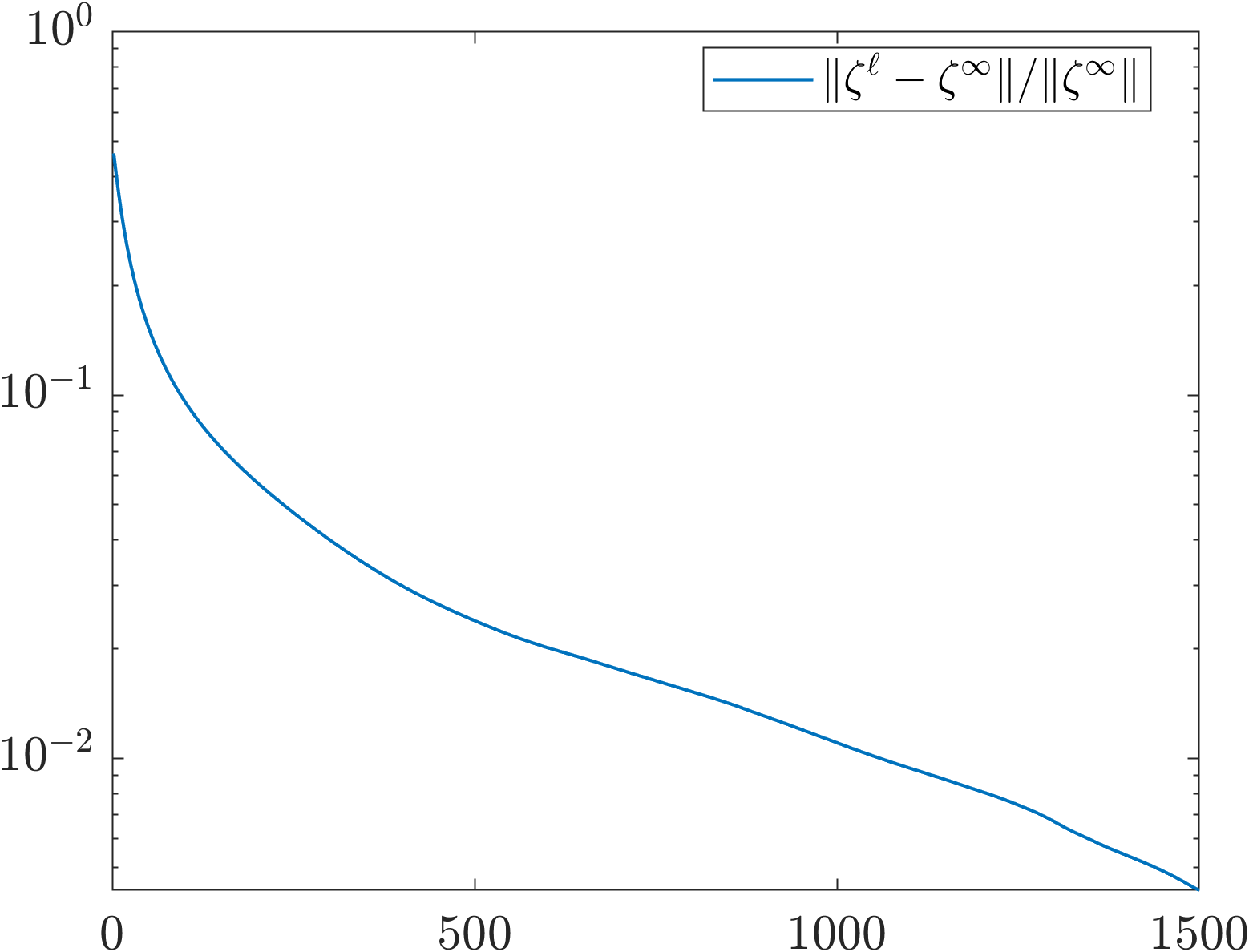}\\
     (a)&(b) & (c) & (d)\\
     \end{tabular}}
     \caption{Decay of the objective value along 500 iterations for \emph{Simu1} (a) and \emph{Simu2} (b). We considered ten random sampling for $p^0$ and $\beta^0$. The continuous line in the plot represents the mean objective value at each iteration, and the shaded area, highlighted in the zoomed region at the centre spanning over 20 iterations, corresponds to the confidence interval related to the standard deviation. Logarithmic plot of the relative distance from the iterates $\zeta^\ell$ to the solution $\zeta^{\infty}$ over 1500 iterations for \textit{Simu1} (c) and \textit{Simu2}~(d).}
     \label{fig:simu1_and_simu2:cost}
\end{figure*}


Eventually, \autoref{fig:simu1_and_simu2:cost} (a)-(b) show the evolution of the mean value of the cost function for both \emph{Simu1} (a) and \emph{Simu2} (b) along 500 iterations for ten different sampling of $p^0$ and $\beta^0$, while \autoref{fig:simu1_and_simu2:cost} (c)-(d)illustrate on a logarithmic scale the relative distance from the iterates to the solution $\|\zeta^\ell - \zeta^\infty\|/\|\zeta^\infty\|$ for \emph{Simu1} (c) and \emph{Simu2} (d), showing the convergence of our algorithm.\\

{   
Additional experiments can be found in Supplementary Material: SM1, showing that for standard wavelet-based image restoration problems
the proposed regularisation outperforms other sparsity measures}.\\ 

 \section{Conclusions}\label{sec:conclusions}
 We investigated a novel approach for the joint reconstruction/feature extraction problem. The novelty in this work lies both in the problem formulation and in the resolution procedure. 
 Firstly, we proposed a new variational model in which we introduced a flexible sparse regularisation term for the reconstruction task. 
 Secondly, we designed {   an inexact version of a TITAN-based block alternating optimisation scheme,} whose aim is to exploit the structure of the problem and the properties of the functions involved in it. We established convergence results for the proposed algorithm whose scope goes beyond the image processing problems considered in our work.
We illustrated the validity of the approach on
  numerical examples in the case of a joint deconvolution-segmentation problem. We also included comparisons with state-of-the-art methods with respect to which our proposal registers a similar qualitative and quantitative performance. An attractive aspect of the proposed work is that the space variant parameters defining the flexible sparse regularisation do not need to be defined in advance, but are inherently estimated by the iterative optimisation procedure. 
  For what concerns the tuning of the hyperparamters of the model, the design of an automatic strategy could be an interesting development of the work, for instance through supervised learning.
}
 \bmhead{Acknowledgments} This project has received funding from the European Un\-ion’s Horizon 2020 research and innovation programme under the Marie Skłodowska-Curie grant agreement No 861137.
The authors thank Ségolène Martin for her careful reading of the initial version of this manuscript.

 

\bibliography{references}

\ifarXiv
    

\section*{Supplementary material}

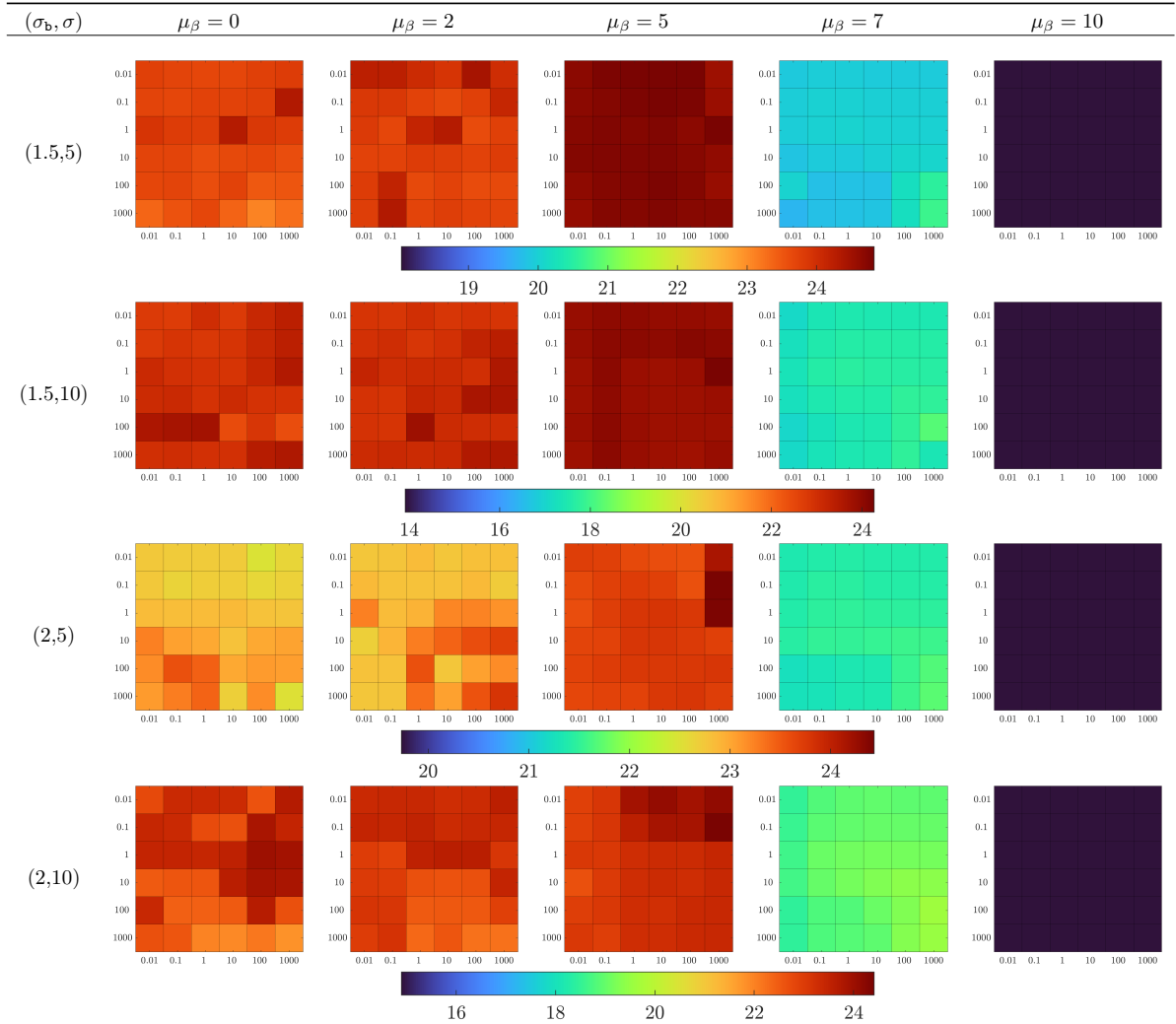

**Fig. 5**: Illustration of the PSNR scores obtained with a grid search on parameters $(\lambda, \zeta)$ for *Westconcord* for 5 different choices for parameter $\mu_\beta$. In each matrix, rows correspond to the values for $\lambda$ and columns to the values for $\zeta$. For each parameter we chose 6 values in a logarithmic scale between $10^{-2}$ and $10^3$.

6

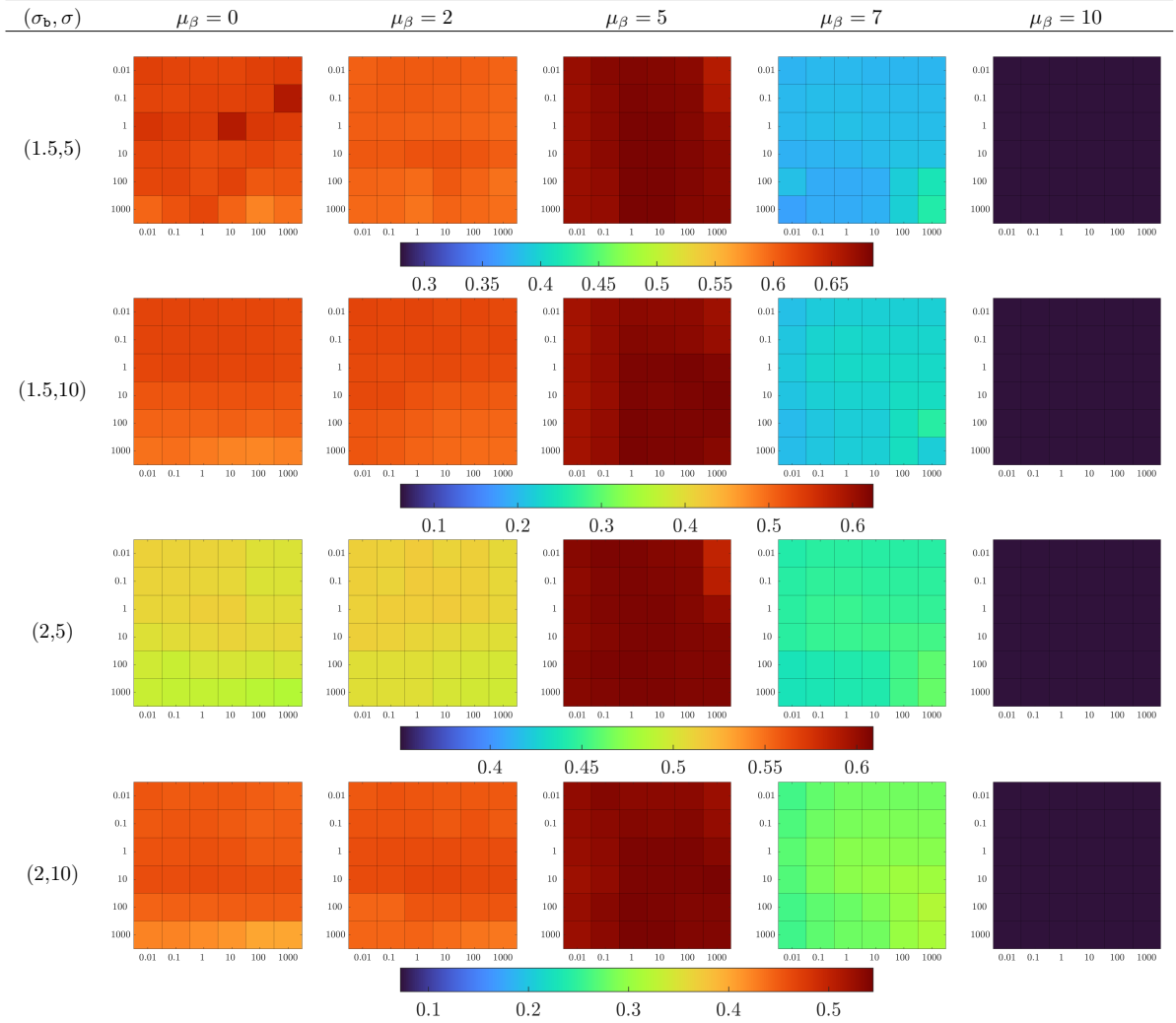

**Fig. 6**: Illustration of the SSIM scores obtained with a grid search on parameters $(\lambda, \zeta)$ for *Westconcord* for 5 different choices for parameter $\mu_\beta$. In each matrix, rows correspond to the values for $\lambda$ and columns to the values for $\zeta$. For each parameter we chose 6 values in a logarithmic scale between $10^{-2}$ and $10^3$.

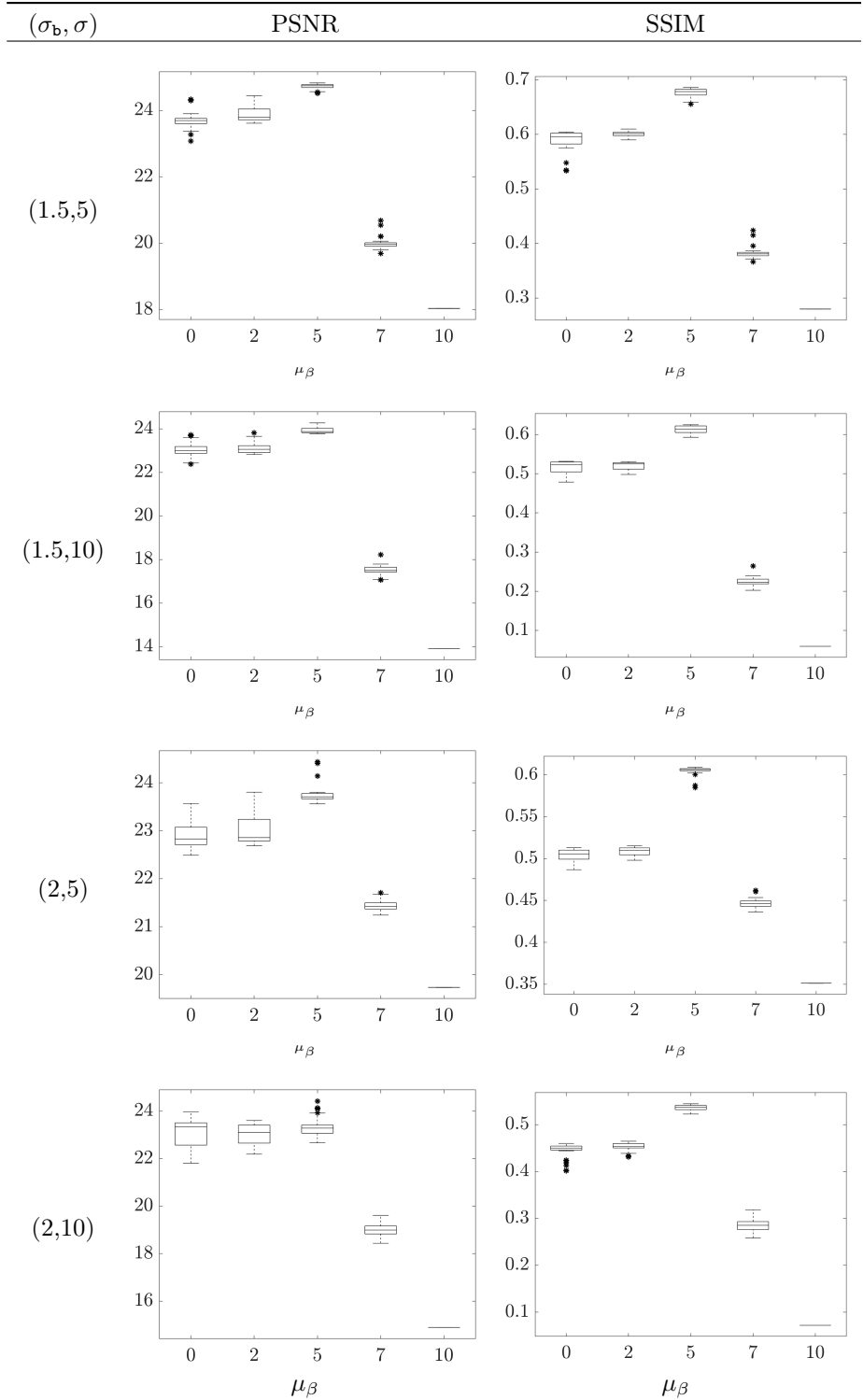

**Fig. 7**: Boxplots of the PSNR scores (left column) and SSIM scores (right column) obtained with a grid search on parameters $(\lambda, \zeta)$ for *Westconcord* for 5 different choices for parameter $\mu_\beta$. The rows corresponds to different combinations of $(\sigma_{\mathtt{b}}, \sigma)$ with $\sigma_{\mathtt{b}} \in \{1.5, 2\}$ and $\sigma \in \{5, 10\}$.

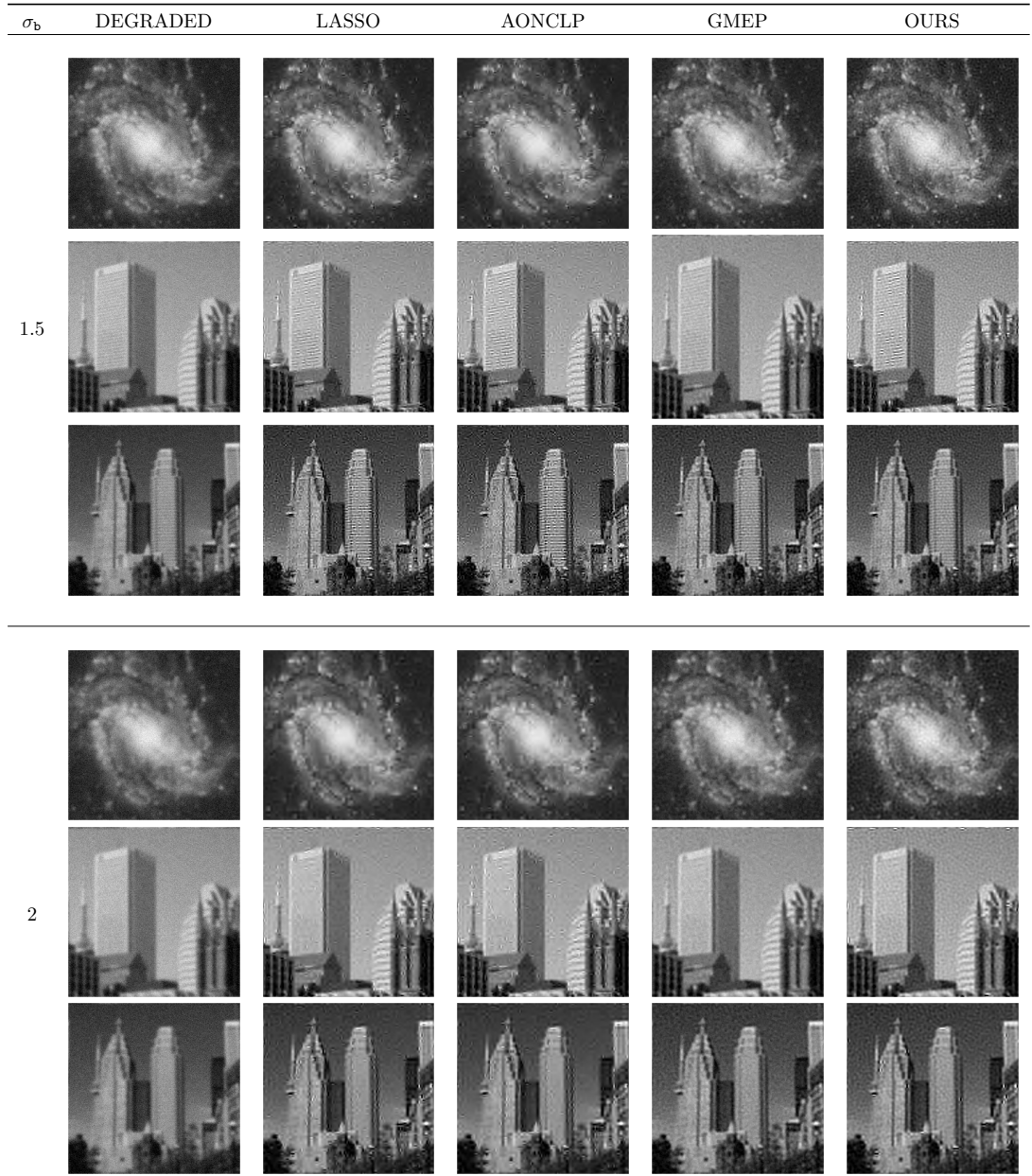

**Fig. 8**: Estimated solutions for *Galaxy*, *Buildings* and *Skyscrapers* and $\sigma = 10$. First column: Degraded image. Second column: LASSO regularisation. Third column: AONCLP regularisation. Fourth column: GMEP regularisation. Fifth column: Proposed regularisation. First line: $\sigma_{\mathtt{b}} = 1.5$, Second line: $\sigma_{\mathtt{b}} = 2$

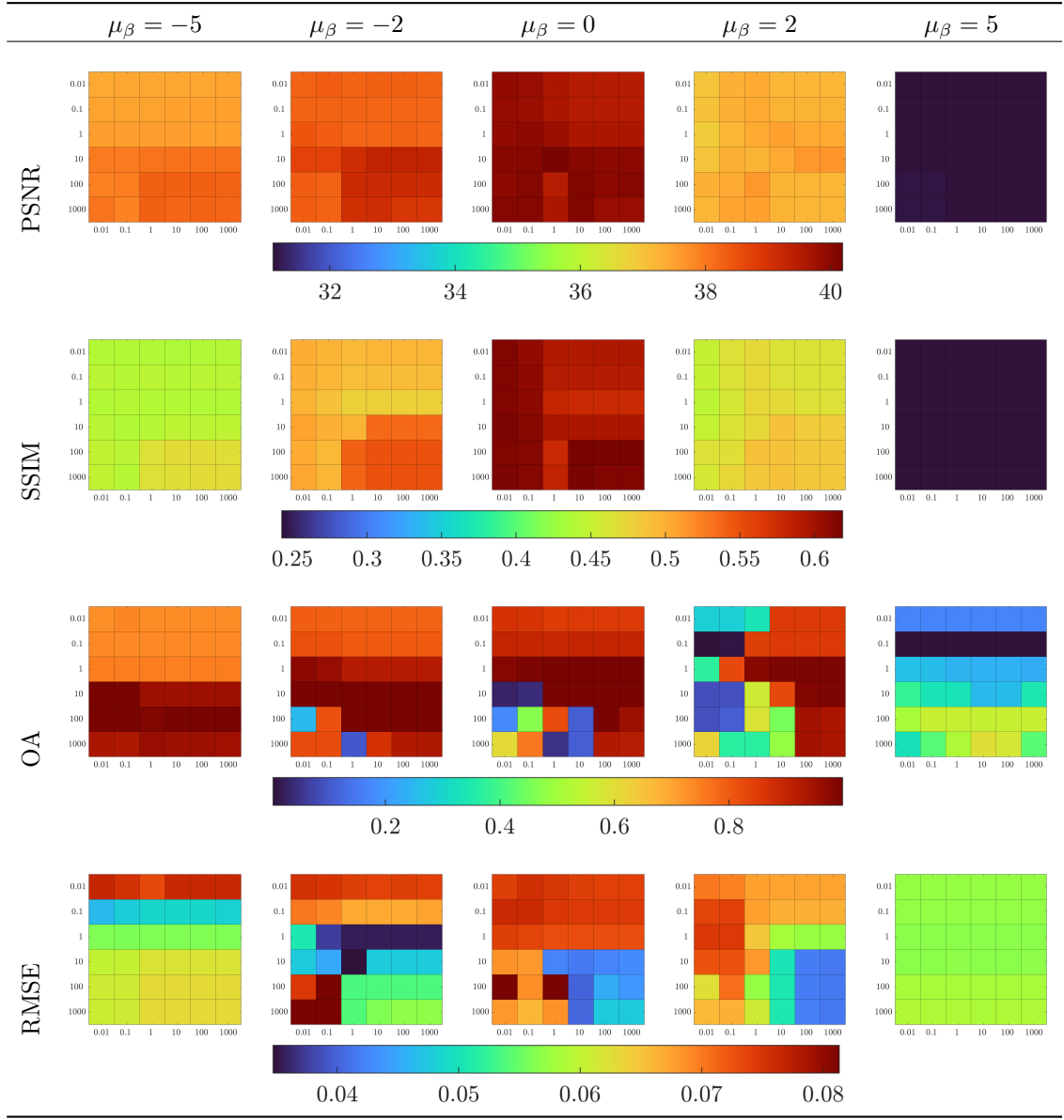

**Fig. 9**: Illustration of the PSNR, SSIM, OA and RMSE values obtained with a grid search on parameters $(\lambda, \zeta)$ for *Simu1* and 5 different choices for parameter $\mu_\beta$. In each matrix, rows correspond to the values for $\lambda$ and columns to the values for $\zeta$. For each parameter we chose 6 values in a logarithmic scale between $10^{-2}$ and $10^3$.

10

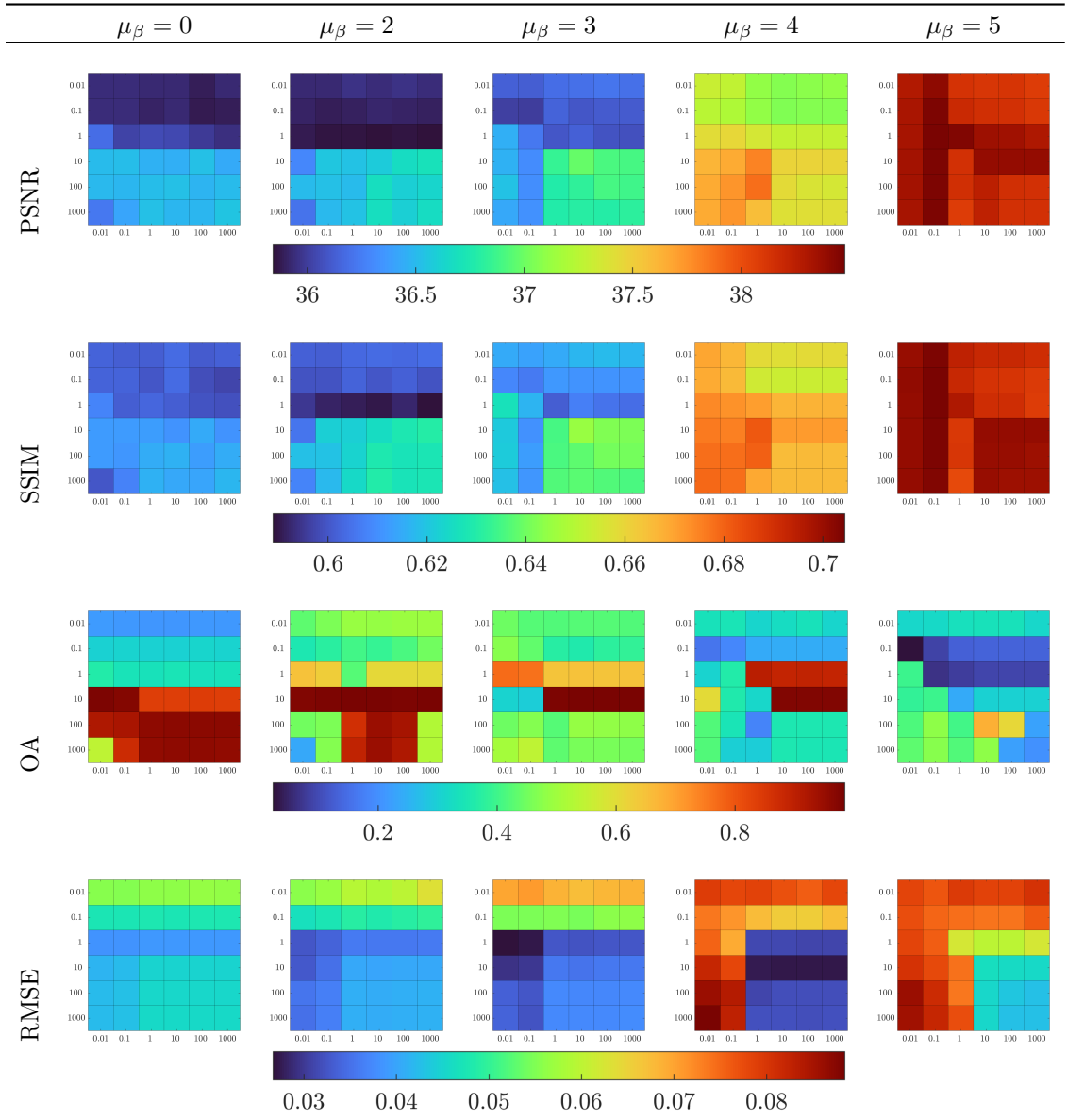

**Fig. 10**: Illustration of the PSNR, SSIM, OA and RMSE values obtained with a grid search on parameters $(\lambda, \zeta)$ for *Simu2* and 5 different choices for parameter $\mu_\beta$. In each matrix, rows correspond to the values for $\lambda$ and columns to the values for $\zeta$. For each parameter we chose 6 values in a logarithmic scale between $10^{-2}$ and $10^3$.

11

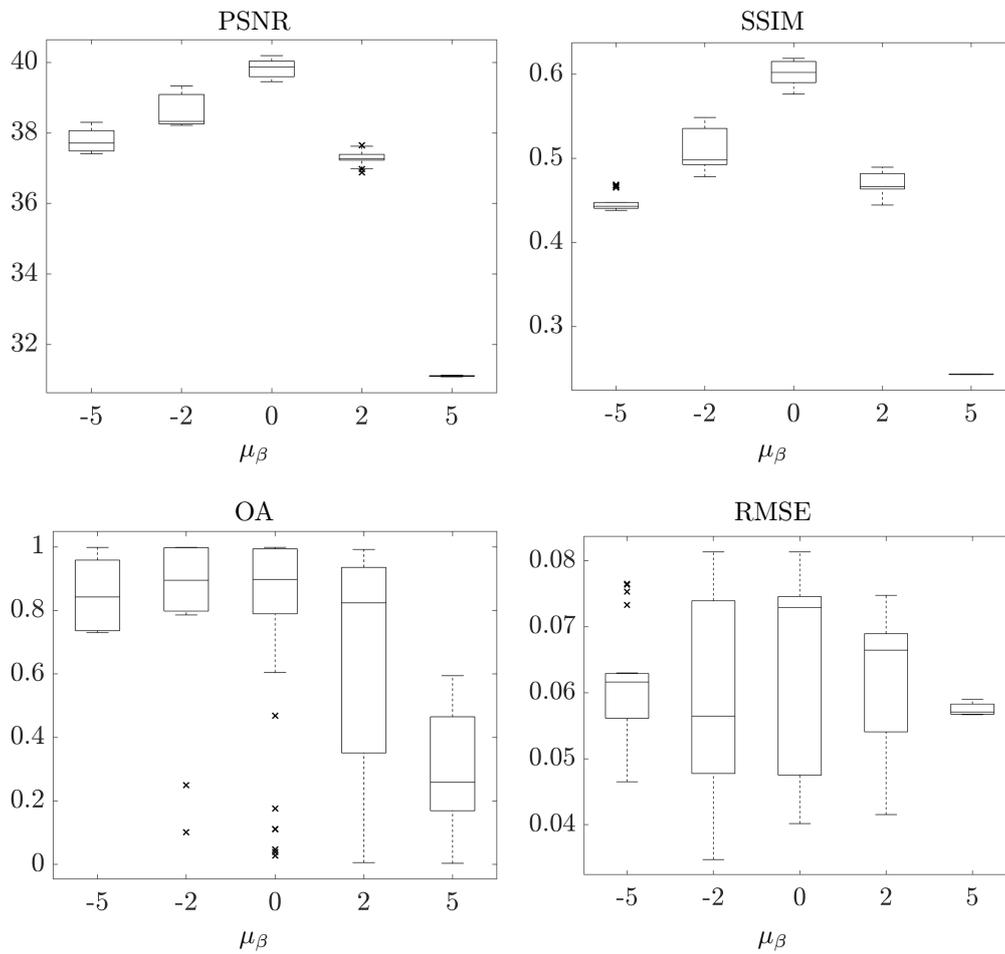

**Fig. 11**: Boxplots of the PSNR, SSIM, OA and RMSE values obtained with a grid search on parameters $(\lambda, \zeta)$ for *Simu1* and 5 different choices for parameter $\mu_\beta$.

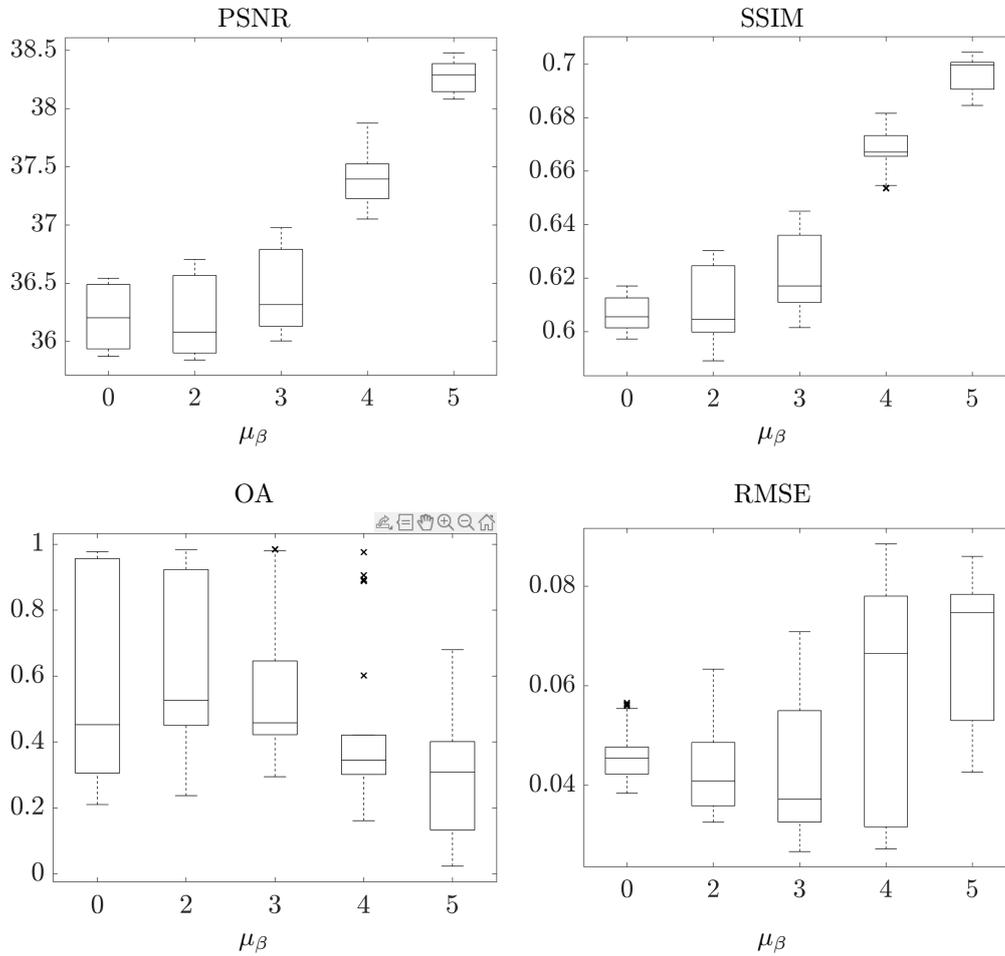

**Fig. 12**: Boxplots of the PSNR, SSIM, OA and RMSE values obtained with a grid search on parameters $(\lambda, \zeta)$ for *Simu2* and 5 different choices for parameter $\mu_\beta$.


\fi
\end{document}